\documentclass[lettersize,journal]{old.IEEEtran}

\usepackage{color}
\usepackage{caption}
\usepackage{times}
\usepackage{helvet}
\usepackage{courier}
\usepackage{amsmath}
\usepackage{amssymb}
\usepackage{amsthm}
\usepackage{graphicx}
\usepackage{array}
\usepackage{booktabs}
\usepackage{amsopn}
\usepackage{amsmath,bm}
\usepackage{algorithm}
\usepackage{algorithmic}
\usepackage{enumerate}
\usepackage{color}
\usepackage{multirow}
\usepackage{bbding}
\usepackage{ragged2e}
\usepackage{threeparttable}
\usepackage{threeparttable} 
\usepackage{dsfont} 
\usepackage{makecell}
\usepackage{url}            
\usepackage{multirow}
\usepackage{microtype}
\usepackage{cite}
\usepackage{outlines}

\usepackage{xcolor,colortbl}
\definecolor{tpamiblue}{rgb}{0.13333333333333333, 0, 1}
\usepackage[colorlinks, linkcolor=tpamiblue, citecolor=tpamiblue]{hyperref}
\usepackage[numbers,sort&compress]{natbib}

\usepackage[capitalize]{cleveref}
\crefname{section}{Sec.}{Secs.}
\Crefname{section}{Section}{Sections}

\crefname{figure}{Fig.}{Figs.}
\Crefname{figure}{Figure}{Figures}

\crefname{table}{Tab.}{Tabs.}
\Crefname{table}{Table}{Tables}

\crefname{algorithm}{Algo.}{Algos.}
\Crefname{algorithm}{Algorithm}{Algorithms}

\usepackage{tikz}
\usetikzlibrary{shadows}
\usetikzlibrary{trees,positioning,shapes,shadows,arrows}
\usepackage[edges]{forest}
\definecolor{hiddendraw}{RGB}{205, 44, 36}

\newcommand{\etal}{\emph{et al.}}

\graphicspath{{figs/}}
\definecolor{blue1}{RGB}{3, 4, 94}
\definecolor{blue2}{RGB}{2, 62, 138}
\definecolor{blue3}{RGB}{0, 119, 188}
\definecolor{blue4}{RGB}{0, 150, 199}

\definecolor{custommagenta}{RGB}{219, 79, 150}
\definecolor{custompurple}{RGB}{112, 81, 211}
\definecolor{customblue}{RGB}{80, 96, 221}

\newcommand{\orcid}[1]{\,\href{https://orcid.org/#1}{\protect\includegraphics[width=8pt]{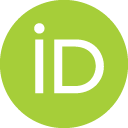}}}

\definecolor{darkblue}{rgb}{0, 0.08235, 0.43921}

\begin{document}

\title{Neural Fields in Robotics: A Survey}

\author{Muhammad Zubair Irshad$^{1\orcid{0000-0002-1955-6194}}$, Mauro Comi$^{2\orcid{0000-0001-7040-5618}}$, Yen-Chen Lin$^{3\orcid{0009-0000-9082-8079}}$, Nick Heppert$^{4\orcid{0000-0002-4347-2644}}$, Abhinav Valada$^{4\orcid{0000-0003-4710-3114}}$ \\ Rares Ambrus$^{1\orcid{0000-0002-3111-3812}}$, Zsolt Kira$^{5\orcid{0000-0002-2626-2004}}$, Jonathan Tremblay$^{3\orcid{0009-0008-4183-6433}}$ \\

\small{$^1$Toyota Research Institute}, {$^2$University of Bristol}, {$^3$Nvidia}, {$^4$University of Freiburg}, {$^5$Georgia Institute of Technology}}

\twocolumn[{%
\renewcommand\twocolumn[1][]{#1}%
\maketitle

\begin{center}
    \vspace{-0.9cm}
    
    \includegraphics[width=1.0\textwidth]{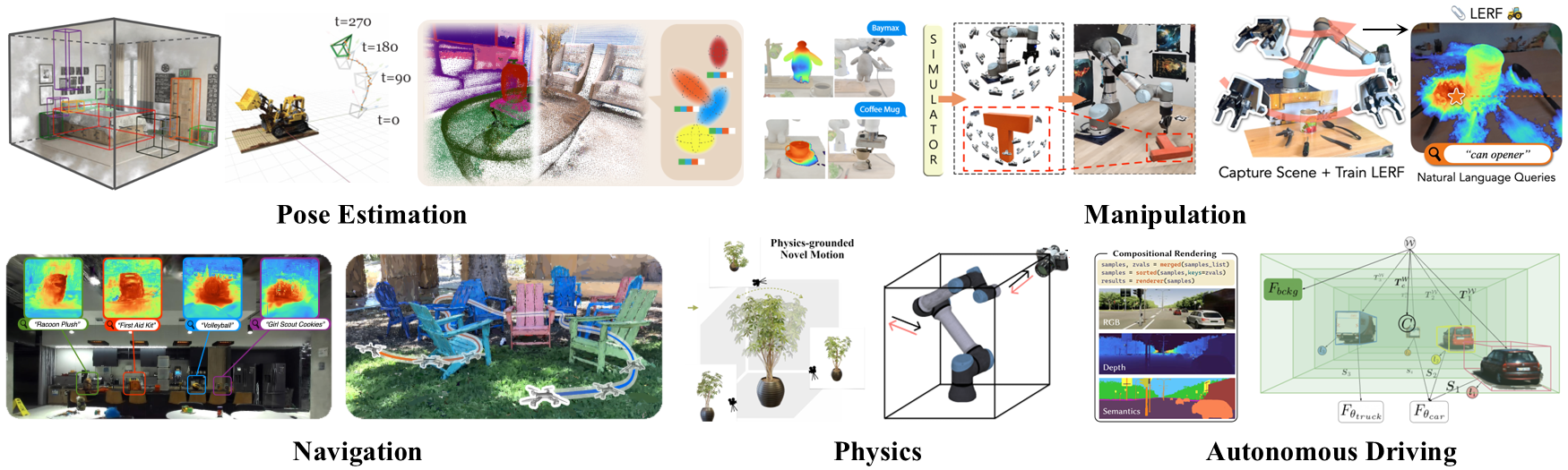}
    \captionsetup{width=\linewidth}
    \captionof{figure}{\textbf{Overview:} This survey paper discusses a large variety of state-of-the-art Neural Field methods that enable robotics applications in pose estimation, manipulation, navigation, physics, and autonomous driving. Images adapted from~\cite{hu2023nerf, yen2020inerf, li2024sgs, shen2023F3RM, lerftogo2023, qureshi2024splatsimzeroshotsim2realtransfer, yu2024legs, chen2024safer, xie2023physgaussian, liu2024differentiable, wu2023mars, ost2021neural}.
    }
   \label{fig:teaser}
\end{center}%
}]

\begin{abstract}
Neural Fields have emerged as a transformative approach for 3D scene representation in computer vision and robotics, enabling accurate inference of geometry, 3D semantics, and dynamics from posed 2D data. Leveraging differentiable rendering, Neural Fields encompass both continuous implicit and explicit neural representations enabling high-fidelity 3D reconstruction, integration of multi-modal sensor data, and generation of novel viewpoints. This survey explores their applications in robotics, emphasizing their potential to enhance perception, planning, and control. Their compactness, memory efficiency, and differentiability, along with seamless integration with foundation and generative models, make them ideal for real-time applications, improving robot adaptability and decision-making. This paper provides a thorough review of Neural Fields in robotics, categorizing applications across various domains and evaluating their strengths and limitations, based on over 200 papers. First, we present four key Neural Fields frameworks: Occupancy Networks, Signed Distance Fields, Neural Radiance Fields, and Gaussian Splatting. Second, we detail Neural Fields' applications in five major robotics domains: pose estimation, manipulation, navigation, physics, and autonomous driving, highlighting key works and discussing takeaways and open challenges. Finally, we outline the current limitations of Neural Fields in robotics and propose promising directions for future research. Project page:
\href{https://robonerf.github.io}{\textit{robonerf.github.io}}
\end{abstract}

\section{Introduction}\label{Sec:Introduction}
\looseness=-1
\IEEEPARstart{R}{\lowercase{obots}} depend on precise and compact representations of their environment to perform a wide array of tasks, from navigating busy warehouses to organizing cluttered homes or assisting in high-stakes search-and-rescue missions. At the core of a typical robotic pipeline is the synergy between perception and action. The perception system gathers sensory data from devices such as RGB cameras, LiDAR, and depth sensors and transforms them into a coherent model of the environment --- such as a 3D map that enables the robot to maneuver through dynamic, obstacle-rich spaces. The quality of this representation directly impacts the robot’s decision-making or policy, which translates the perceived environment into actions, enabling it to avoid moving forklifts, pick up scattered objects, or plan a safe path in an emergency.
\looseness=-1
Traditionally, robots have modeled their environments using data structures like point clouds~\cite{fan2017point, irshad2022centersnap, achlioptas2018learning}, voxel grids~\cite{zhou2018voxelnet}, meshes~\cite{wang2018pixel2mesh, chen2020bsp, liao2018deep}, and Truncated Signed Distance Functions (TSDF)~\cite{breyer2020volumetric}. While these representations have advanced robotic capabilities, they struggle to capture fine geometric details, particularly in complex or dynamic environments, leading to suboptimal performance in adaptable scenarios.

To overcome these limitations, Neural Fields (NFs)~\cite{xie2021neural} have emerged as a promising alternative, offering continuous, differentiable mappings from spatial coordinates to physical quantities like color or signed distance. Unlike traditional data structures, NFs can model 3D environments as continuous functions parameterized by neural networks or Gaussian distributions. This enables them to represent complex geometries and fine details more efficiently~\cite{mildenhall2020nerf, park2019deepsdf}. NFs can be optimized using gradient-based methods with various types of real-world sensory data, including images and depth maps, to produce high-quality 3D reconstructions. In the realm of robotics, NFs provide several distinct advantages over traditional methods:

\begin{itemize} \item \textbf{High-Quality 3D Reconstructions:} NFs generate detailed 3D representations of environments, which are crucial for tasks like navigation, manipulation, and scene understanding~\cite{wang2021neus, muller2022instant,  guedon2024sugar, refine, takikawa2021nglod}. \item \textbf{Multi-Sensor Integration:} NFs can seamlessly integrate data from multiple sensors, such as LiDAR and RGB cameras, providing a more robust and adaptable perception of the environment~\cite{Huang_2023_ICCV, Hwang_2023_WACV}. \item \textbf{Continuous and Compact Representations:} Unlike voxel grids or point clouds, which are inherently discrete, NFs offer continuous representations that capture fine spatial details using fewer parameters, enhancing computational efficiency~\cite{NEURIPS2020_55053683, mildenhall2020nerf}. \item \textbf{Generalization and Adaptation:} Once trained, NFs can generate novel viewpoints of a scene, even from previously unseen perspectives, which is particularly valuable for exploration or manipulation tasks. This ability is enabled by generalizable NeRF methods~\cite{yu2020pixelnerf, wu2023reconfusion, zeronvs}. \item \textbf{Integration with Foundation Models:} NFs can be combined with foundation models like CLIP~\cite{radford2021learning} or DINO~\cite{oquab2023dinov2}, enabling robots to interpret and respond to natural language queries or other semantic inputs~\cite{kobayashi2022decomposing, lerf2023}. \end{itemize}

Recent advances in generative AI~\cite{ho2020denoising} have further expanded the capabilities of NFs by leveraging synthetic data as supervisory signals, thereby reducing reliance on real-world observations. This paradigm shift allows NFs to be optimized in scenarios where real-world data collection is impractical or costly. Importantly, it positions NFs as a key link between generative AI and robotics. While generative priors from 2D data are powerful, they often lack the spatial coherence needed for effective robotic decision-making. NFs integrate these priors with sparse real-world data~\cite{wu2023reconfusion}, enabling them to model sensory and motor spaces in scenarios constrained by physical environments, such as limited sensor configurations and occlusions.

\begin{figure}[t!] 
    \centering
    \includegraphics[width=1.0\columnwidth]{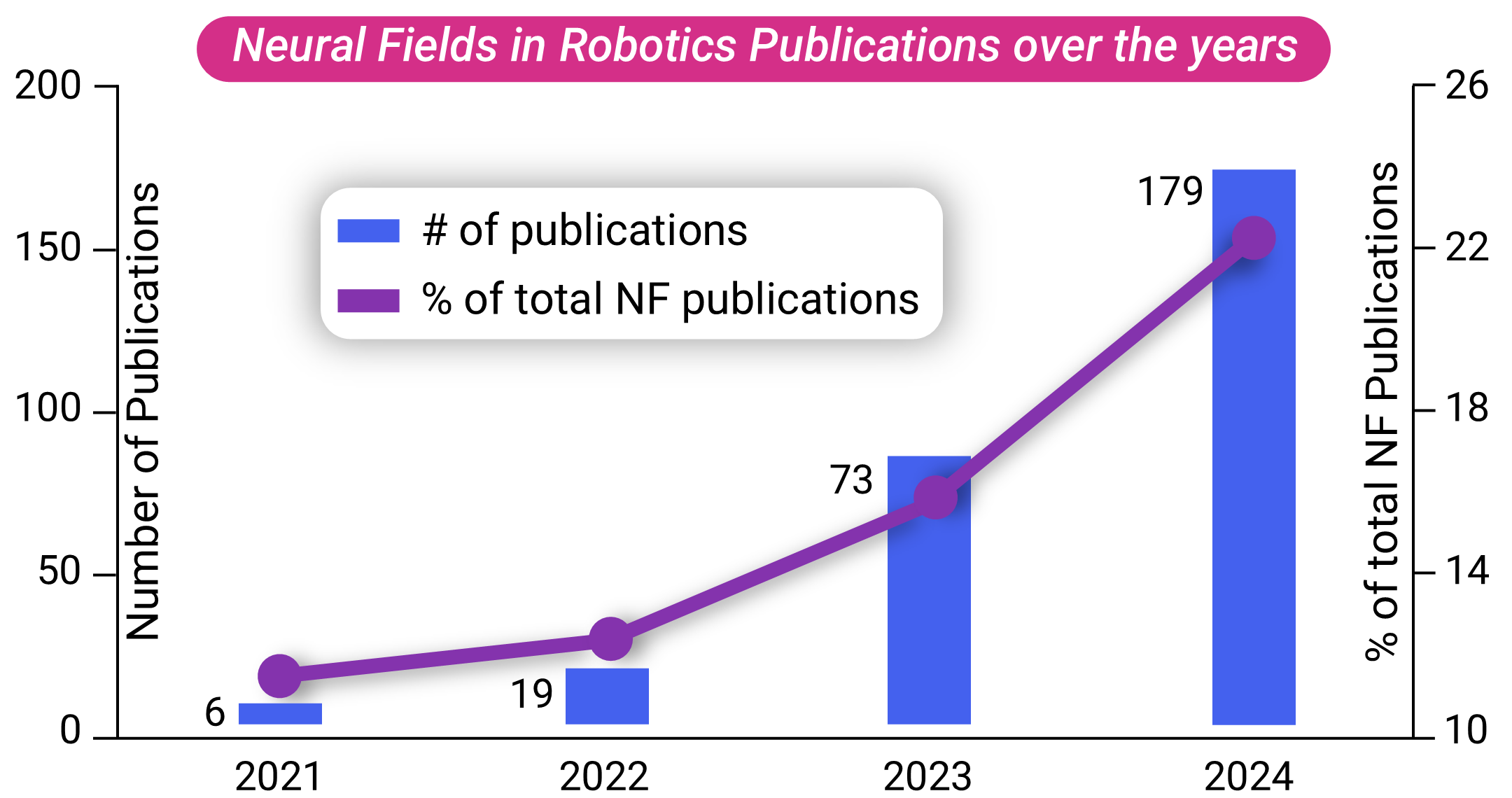}%
    \caption{\textbf{Growth of Neural Fields in Robotics:}  plotted as a rough number of publications vs. \% of total neural field publications per year.}
    \vspace{-0.4cm}
    \label{fig:papers_stats}
\end{figure}

\begin{figure*}[t!] 
    \centering
    \includegraphics[width=1.0\textwidth]{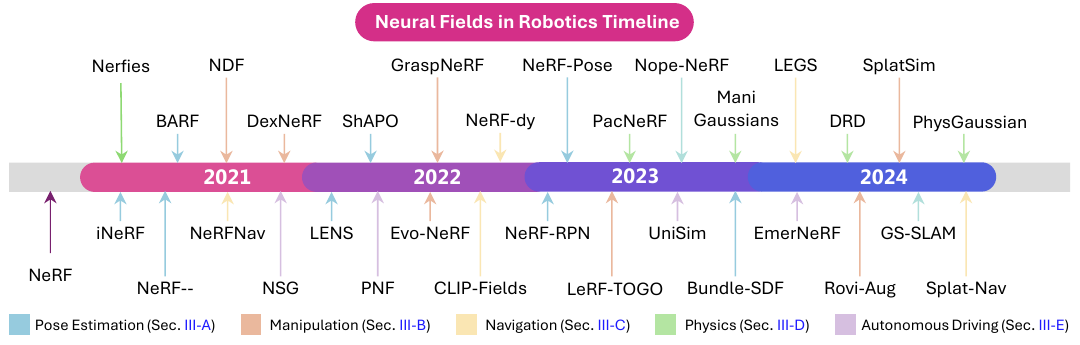}%
    \caption{~\textbf{Timeline} of Neural Fields in Robotics paper showing key papers over the years divided into 5 major application areas.}
    \vspace{-0.5cm}
    \label{fig:timeline}
\end{figure*}

Given these advantages, the application of NFs in robotics is a rapidly growing area of research. Figs.~\ref{fig:teaser} and~\ref{fig:papers_stats} provide an overview of NF applications in robotics and highlight the rise in NF-related robotics publications over time. In this paper, we aim to structure and analyze their impact on the field. The manuscript is organized as follows: Sec.~\ref{sec:theoretical_background} covers the formulation of NFs, while Sec.~\ref{sec:neural_fields_for_robotics} highlights their benefits across various domains, categorized into distinct themes:
\begin{itemize}
    \item \textbf{Pose Estimation}, focuses on NF as a scene or object representation in camera pose estimation, object pose estimation, and Simultaneous Localization and Mapping~(SLAM)~(\cref{sec:nf_pose}).
    \item \textbf{Manipulation}, discusses how NFs' accurate 3D reconstruction assists robots in manipulating objects~(\cref{sec:nf_manipulation}).
    \item \textbf{Navigation}, highlights the role of NFs in enhancing robotic navigation by enabling accurate and efficient perception of real-world environments~(\cref{sec:nf_navigation}).    
    \item \textbf{Physics}, explores how NFs enable robots to reason about physical interactions to improve their understanding of real-world dynamics~(\cref{sec:nf_physics}).    
    \item \textbf{Autonomous Driving}, focuses on NFs' role in building photorealistic simulators for the real world (\cref{sec:nf_ad}).
\end{itemize}

We conclude by discussing several research directions and challenges in Sec.~\ref{sec:future_works}. To the best of our knowledge, this survey represents one of the first comprehensive examinations of Neural Fields in the domain of robotics. We complement the closest concurrent survey~\cite{wang2024nerfinrobotics} that focuses on NeRFs by covering a comprehensive set of fields, including 3DGS, Occupancy, Signed Distance Fields, and beyond. By integrating insights from various dimensions, this survey aims to provide a holistic understanding of the current state of NF in robotics applications, highlighting recent achievements, upcoming challenges, and unexplored areas within robotics.

\section{Formulation of Neural Fields}
\label{sec:theoretical_background}
We begin by defining several types of fields that are key to the formulation of Neural Fields. In mathematics and physics, a field is a descriptor of a quantity with a value assigned to every point in space and time. Formally, this is expressed as:

\begin{itemize}
  \item \textbf{Scalar fields}: A scalar field $f : \mathbb{R}^n \rightarrow \mathbb{R}$ assigns a scalar value ({\em e.g.} temperature, pressure) to every point in an $n$-dimensional space. Mathematically, it is defined as $f(x, y, z)$ for a 3D space, where $(x, y, z)$ are the coordinates in space and $f$ is the scalar value at that point. Examples of scalar fields are the Occupancy Fields (\cref{subsec:occupancy}) and Signed Distance Fields (\cref{subsec:sdf}).

  \item \textbf{Vector fields}: A vector field $F : \mathbb{R}^n \rightarrow \mathbb{R}^m$ is an extension of a scalar field that associates a vector ({\em e.g.}, velocity or force) with every point in space. For example, in three dimensions, a possible formulation is given by $F(x, y, z) = (F_0(x, y, z), ..., F_{m-1}(x, y, z))$, where each element $F_i$ represents the vector values corresponding to the input coordinate $(x, y, z)$. This function can either be represented as a neural network~(\cref{sec:nerf}),
   a volumetric grid~\cite{sun2022direct}, or the mix of both~\cite{muller2022instant}.
\end{itemize}

\subsection{Occupancy Fields} \label{subsec:occupancy}
Occupancy represents the binary state of whether a point $\mathbf{p}$ in space is occupied by a surface $S$ or not:
\vspace*{-\abovedisplayskip}
\[ o(\mathbf{p}) = \begin{cases} 0 & \text{if } \mathbf{p} \text{ is inside } S, \\1 & \text{if } \mathbf{p} \text{ is outside } S. \end{cases} \]
\vspace*{-\abovedisplayskip}

This idea can be extended to the continuous case, where occupancy is represented by the probability $o(\mathbf{p}) \in \left[0, 1\right]$ indicating the likelihood of $\mathbf{p}$ being inside or outside the surface. Occupancy Networks~\cite{mescheder2019occupancy} leverage this idea by learning a continuous function, or NF, which maps points $\mathbf{p} \in \mathbb{R}^n$ to occupancy probabilities, conditioned on input observations $\mathbf{x}$ ({\em e.g.}, point clouds or images). This function, $f_\theta(\mathbf{p} ~|~ \mathbf{x})$, is optimized during training by sampling points $\mathbf{p}_i$ and minimizing the cross-entropy loss between predicted occupancy $f_\theta(\mathbf{p}_i ~|~ \mathbf{x})$ and ground truth occupancy $o(\mathbf{p}_i, \mathbf{x})$. This approach enables smooth, continuous surface representations, which can later be thresholded to recover discrete occupancy.

\subsection{Signed Distance Fields (SDFs)}\label{subsec:sdf}
Signed Distance Fields assign each point in space the shortest distance to the surface boundary of an object, with the sign indicating whether the point is inside or outside. For a point $\mathbf{p}$ and surface $S$, the SDF $d(\mathbf{p})$ is defined as:
\vspace*{-\abovedisplayskip}
\[ d(\mathbf{p}) = \begin{cases} -\min_{\mathbf{q} \in S} \|\mathbf{p} - \mathbf{q}\| & \text{if } \mathbf{p} \text{ is inside } S, \\ \min_{\mathbf{q} \in S} \|\mathbf{p} - \mathbf{q}\| & \text{if } \mathbf{p} \text{ is outside } S, \end{cases} \]
\vspace*{-\abovedisplayskip}

\noindent where $\|\mathbf{p} - \mathbf{q}\|$ is the Euclidean distance between points $\mathbf{p}$ and $\mathbf{q}$. When representing the SDF as an NF, during training, points $\mathbf{p}_i$ are sampled, and their distance to the closest surface provides the supervision signal \cite{park2019deepsdf}.

\subsection{Radiance Fields} 
\looseness=-1
Radiance Fields represent the distribution of light in 3D space, associating each point with a radiance value (light intensity and color) in every direction. This can be described using a function $L : \mathbb{R}^3 \times S^2 \rightarrow \mathbb{R}^3$, where $L(p, \omega)$ gives the light radiance at point $p$ in direction~$\omega$, with~$p$ in 3D space and~\textbf{}$\omega$ on the unit sphere $S^2$ representing all possible directions.

\subsubsection{Neural Radiance Fields~(NeRF)}
\label{sec:nerf}
NeRFs~\cite{mildenhall2020nerf} represent scenes as volumetric fields of density~$\sigma$ and RGB color~$\mathbf{c}$ using a neural network. The weights of NeRFs are optimized per scene using input RGB images, and their camera poses. After training, the density field captures scene geometry, while the color field models the view-dependent appearance. A multilayer perceptron (MLP) with parameters $\Theta$ predicts the density $\sigma$ and RGB color $\mathbf{c}$ for each point based on its 3D position $\mathbf{x} = (x, y, z)$ and viewing direction $\mathbf{d}$. To address the spectral bias of neural networks in low-dimensional spaces~\cite{tancik2020fourfeat}, positional encoding $\gamma(\cdot)$ is applied to inputs: $(\sigma, \mathbf{c}) \leftarrow F_{\Theta}(\gamma(\mathbf{x}), \gamma(\mathbf{d}))$.

To render a pixel, a camera ray $\mathbf{r}(t) = \mathbf{o} + t\mathbf{d}$ is cast from the camera center $\mathbf{o}$ in direction $\mathbf{d}$. $K$ points $\{\mathbf{x}_k = \mathbf{r}(t_k)\}_{k=1}^K$ are sampled along the ray and passed through the MLP to generate densities and colors $\{\sigma_k, \mathbf{c}_k\}_{k=1}^K$. These are then combined to estimate the pixel color $\hat{\mathbf{C}}(r)$ via volume rendering~\cite{kajiya84}, using a numerical quadrature approximation~\cite{max1995optical}
\newcommand{\bigT}{T}
\vspace*{-\abovedisplayskip}
\begin{equation} \label{eq:volume_rendering}
    \hat{\textbf{C}}(\mathbf{r}) = \sum_{k=1}^{K} \bigT_k \bigg(1- \exp\Big(-\sigma_k (t_{k+1} - t_k)\Big)\bigg) \textbf{c}_k, \\ \text{with} 
\end{equation}
where $\bigT_k = \text{exp} \Big(-\sum_{k' < k} \sigma_{k'} (t_{k'+1} - t_{k'})\Big)$ can be interpreted as the probability that the ray successfully transmits to point $\mathbf{r}(t_k)$. NeRF is trained by minimizing the photometric loss, $\mathcal{L}_{\text{photo}} = \sum_{\mathbf{r} \in \mathcal{R}} \left\| \hat{\mathbf{C}}(\mathbf{r}) - \mathbf{C}(\mathbf{r}) \right\|_2^2$, where $\mathbf{C}(\mathbf{r})$ is the observed RGB value for ray $\mathbf{r}$ in a sampled set of rays $\mathcal{R}$.

While vanilla NeRF achieves photorealistic results, it is time-consuming to train and render from a pretrained NeRF. To reduce these costs, several improvements are proposed: \textit{a)}~using encodings with better speed/quality trade-offs~\cite{mueller2022instant}, \textit{b)}~adopting smaller neural networks to reduce memory-bandwidth demands~\cite{Fridovich-Keil_2022_CVPR, sun2022direct}, and \textit{c)}~skipping ray marching steps in empty space to cut down the computational cost of neural volume rendering~\cite{li2022nerfacc}. These optimizations accelerate NeRF training and inference by several orders of magnitude, enabling real-time use in time-sensitive applications.

\begin{figure*}[t!] 
    \centering
    \includegraphics[width=1.0\textwidth]{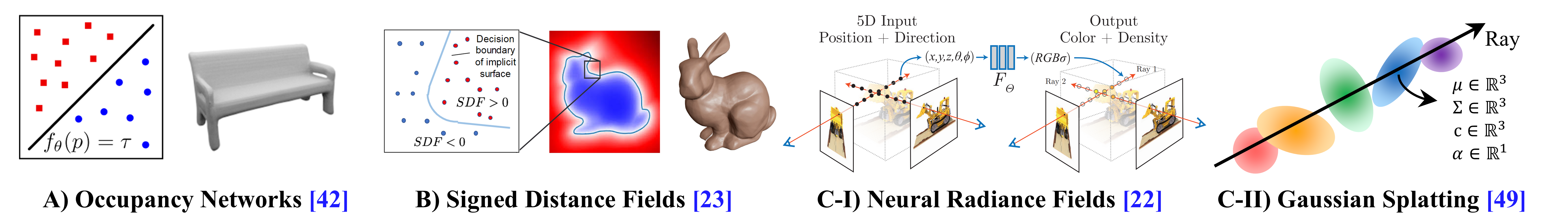}%
    \caption{\textbf{Neural Field Representations:} Section~\ref{sec:theoretical_background} discusses four core Neural Field representations --- Occupancy Networks~\cite{mescheder2019occupancy}, Signed Distance Fields~\cite{park2019deepsdf}, Neural Radiance Fields~\cite{mildenhall2020nerf}, and 3D Gaussian Splatting~\cite{kerbl20233d}.}
    \vspace{-0.4cm}
    \label{fig:nf_representations}
\end{figure*}

\subsubsection{3D Gaussian Splatting} \label{sec:3dgs}
3D Gaussian Splatting~\cite{kerbl20233d} represents a scene as a collection of anisotropic 3D Gaussians, each defined by a mean vector $\boldsymbol{\mu} \in \mathbb{R}^3$, covariance matrix $\mathbf{\Sigma} \in \mathbb{R}^3$, color vector $\mathbf{c} \in \mathbb{R}^3$, and opacity scalar $\alpha$. The influence of a Gaussian on a point $\mathbf{x} \in \mathbb{R}^3$ is given by:
\vspace*{-\abovedisplayskip}
\begin{equation} \label{eq:gaussian_influence}
f(\mathbf{x};\mathbf{\Sigma}, \boldsymbol{\mu}) = \exp(-\frac{1}{2} (\mathbf{x} - \boldsymbol{\mu})^T \mathbf{\Sigma}^{-1} (\mathbf{x} - \boldsymbol{\mu})),
\end{equation}

which quantifies each Gaussian's contribution based on spatial proximity. To render the scene onto a 2D plane, the 3D covariance $\mathbf{\Sigma}$ is projected into 2D as $\mathbf{\Sigma'} = \mathbf{J} \mathbf{W} \mathbf{\Sigma} \mathbf{W}^T \mathbf{J^T}$~\cite{zwicker2001surface}, where $\mathbf{W}$ is the projective transformation and $\mathbf{J}$ is the Jacobian. The 2D mean vector $\boldsymbol{\mu}'$ is obtained via perspective projection $\text{Proj}(\boldsymbol{\mu} | \mathbf{E}, \mathbf{K})$, using the camera's extrinsic $\mathbf{E}$ and intrinsic $\mathbf{K}$ matrices. To ensure valid optimization of 3D covariance $\mathbf{\Sigma}$, it is decomposed as $\mathbf{\Sigma} = \mathbf{R} \mathbf{S} \mathbf{S}^T \mathbf{R}^T$, where $\mathbf{R}$ and $\mathbf{S}$ are rotation and scaling matrices, respectively. Finally, the color of each pixel $p$ in the image can be calculated as:
\vspace*{-\abovedisplayskip}
\begin{equation} \label{eq:rasterisation}
I(p) = \sum_{i=1}^{N} f(p; \boldsymbol{\mu'}_i, \mathbf{\Sigma'}_i) \ \alpha_{i} \ \mathbf{c}_{i} \prod_{j=1}^{i-1} 1 - f(p; \boldsymbol{\mu'}_{j}, \mathbf{\Sigma'}_{j}) \ \alpha_{j}.
\end{equation}
\vspace*{-\abovedisplayskip}

Overall, scalar, vector, and radiance fields, comprising Occupancy networks, Signed Distance Fields, Neural Radiance Fields, and 3D Gaussian Splatting~(as shown in Fig.~\ref{fig:nf_representations}) form the foundation of neural representations, enabling them to capture complex geometries and spatial relationships with far greater detail than traditional methods. In the next section, we explore how these mathematical tools unlock a variety of robotics applications, including pose estimation, manipulation, navigation, physical property inference, and autonomous driving in challenging environments.
\section{Neural Fields for Robotics}
\label{sec:neural_fields_for_robotics}

\begin{figure*}[htbp]
 \centering
 \vspace{-0.4cm}
\begin{forest}
  forked edges,
  for tree={
    grow=east,
    reversed=true,
    anchor=base west,
    parent anchor=east,
    child anchor=west,
    base=left,
    font=\small,
    rectangle,
    draw={hiddendraw, line width=0.6pt},
    rounded corners,align=left,
    minimum width=2.5em,
    edge={black, line width=0.55pt},
    l+=1.9mm,
    s sep=7pt,
    inner xsep=7pt,
    inner ysep=8pt,
    ver/.style={rotate=90, rectangle, draw=custommagenta, rounded corners=1.5mm, fill=white, text centered,  text=black, child anchor=north, parent anchor=south, anchor=center, font=\fontsize{10}{10}\selectfont, line width=0.5mm},
    level2/.style={rectangle, draw=custompurple, fill=white, 
    text centered, anchor=west, text=black, font=\fontsize{8}{8}\selectfont, text width = 7em, line width=0.45mm},
    level3/.style={rectangle, draw=customblue, fill=white,   fill opacity=1.0,text centered, anchor=west, text=black, font=\fontsize{8}{8}\selectfont, text width = 2.7cm, align=center, line width=0.47mm},
    level3_2/.style={rectangle, draw=none, fill=custompurple,   fill opacity=1.0,text centered, anchor=west, text=black, font=\fontsize{8}{8}\selectfont, text width = 3.0cm, align=center},
    level3_1/.style={rectangle, draw=none, fill=custompurple,   fill opacity=1.0, text centered, anchor=west, text=black, font=\fontsize{8}{8}\selectfont, text width = 3.2cm, align=center},
    level4/.style={rectangle, draw=black, text centered, anchor=west, text=black, font=\fontsize{8}{8}\selectfont, align=center, text width = 2.9cm},
    level5/.style={rectangle, draw=black, text centered, anchor=west, text=black, font=\fontsize{8}{8}\selectfont, align=center, text width = 14.7em},
    level5_1/.style={rectangle, draw=black, text centered, anchor=west, text=black, font=\fontsize{8}{8}\selectfont, align=center, text width = 26.1em},
  },
  where level=1{text width=5em,font=\scriptsize,align=center}{},
  where level=2{text width=6em,font=\tiny,}{},
  where level=3{text width=6em,font=\tiny}{},
  where level=4{text width=5em,font=\tiny}{},
  where level=5{font=\tiny}{},
  [Neural Fields in Robotics, ver
    [Pose Estimation\\(\hypersetup{linkcolor=tpamiblue}\cref{sec:nf_pose}), level2
        [Camera Pose Estimation, level3
                [Lens~\cite{moreau2022lens} (2022){,}
                NeRF$--$~\cite{wang2021nerf} (2022){,} LocalRF~\cite{meuleman2023localrf} (2023) , level5_1]]
        [Object Pose Estimation, level3
                [NeRF-Det~\cite{xu2023nerfdet} (2023){,}
                BundleSDF~\cite{wen2023bundlesdf} (2023) {,} NeRF-loc~\cite{sun2023nerf} (2023) , level5_1]]]        
    [Manipulation \\(\hypersetup{linkcolor=tpamiblue}\cref{sec:nf_manipulation}), level2
        [Occupancy, level3
        [NDF~\cite{simeonov2022neural} (2021){,}
            L-NDF~\cite{chun2023local} (2023){,}
            GIGA~\cite{jiang2021synergies} (2021) ,level5_1]
        ]
        [Radiance Fields, level3
            [Dex-NeRF~\cite{ichnowski2021dex} (2021){,} GaussianGrasper~\cite{zheng2024gaussiangrasper} (2024){,}  Evo-NeRF~\cite{kerr2022evo} (2022){,}\\ SPARTN~\cite{zhou2023nerf} (2023){,} ManiGaussians~\cite{lu2024manigaussian} (2024){,} GraspNeRF~\cite{dai2023graspnerf} (2023),level5_1]
        ]
        [SDF, level3
            [NeuralGrasp~\cite{khargonkar2023neuralgrasps} (2023){,} NGDF~\cite{weng2023neural} (2023){,} VGN~\cite{breyer2021volumetric} (2021), level5_1]
        ]
        [Feature Fields, level3
            [DFF~\cite{shen2023F3RM} (2023){,} LeRF-TOGO~\cite{lerftogo2023} (2023){,} GNFactor~\cite{ze2023gnfactor} (2023) , level5_1]
        ]
        [Visual \& Tactile sensing, level3
            [Tactile-3DG~\cite{comi2024snap} (2024){,} Touch-GS~\cite{swann2024touch} (2024), level5_1]
        ]
        [Diffusion Models, level3
            [SE(3)-Diffusion Fields~\cite{urain2023se3diffusionfieldslearningsmoothcost} (2023){,} Rovi-Aug~\cite{chen2024roviaug} (2024){,} Vista~\cite{tian2024vista} (2024), level5_1]
        ]
    ]
    [Navigation \\
    (\hypersetup{linkcolor=tpamiblue}\cref{sec:nf_navigation}), level2
        [Planning, level3
            [CATNips~\cite{chen2023catnips} (2023){,} NeRF-dy~\cite{li20223d} (2022),level5_1
            ]
        ]
        [Exploration, level3
            [AutoNeRF~\cite{marza2023autonerf} (2023){,} Active Neural Mapping~\cite{Yan2023iccv} (2023), level5_1
            ]
        ]
        [Visual Localization, level3
            [RNR-map~\cite{Kwon_2023_CVPR} (2023){,} NeRF-IBVS~\cite{wang2024nerf} (2024){,} Splat-Nav~\cite{chen2024splat} (2024) , level5_1
            ]
        ]
        [Feature Field, level3
            [Clip-Fields~\cite{mahi2022clip} (2023){,} GaussNav~\cite{lei2024gaussnav} (2024){,} LEGS~\cite{yu2024legs} (2024) , level5_1
            ]
        ]
    ]
    [Physics \\(\hypersetup{linkcolor=tpamiblue}\cref{sec:nf_physics}), level2
        [
            Model-Free, level3
            [
                \mbox{D-NeRF~\cite{pumarola2020d} (2020),} 
                \mbox{Dynamic 3DGS~\cite{luiten2023dynamic} (2023),}
                \mbox{DRD~\cite{liu2024differentiable} (2024)},
                level5_1
            ]
        ]
        [
            Model-Based, level3
            [
                \mbox{PAC-NeRF \cite{li2023pacnerf} (2023),}
                \mbox{PhysGaussian \cite{xie2023physgaussian} (2023)},
                level5_1
            ]
        ]            
    ]
    [Autonomous Driving \\ (\hypersetup{linkcolor=tpamiblue}\cref{sec:nf_ad}), level2
        [Manipulability, level3
            [NeuralSceneGraphs~\cite{ost2021neural} (2021){,} StreetNeRF~\cite{xie2023snerf} (2023){,} PNF~\cite{kundu2022panoptic} (2022) , level5_1
            ]
        ]
        [Photorealistic Simulators, level3
            [MARS~\cite{wu2023mars} (2023){,} UniSim~\cite{yang2023unisim} (2023), level5_1
            ]
        ]
        [Generalizability, level3
            [NeO 360~\cite{irshad2023neo360} (2023){,} Neural Groundplans~\cite{sharma2022neural} (2023){,} 6Imgto3D~\cite{gieruc20246imgto3d} (2024) , level5_1
            ]
        ]
    ]
]
\end{forest}
\caption{\textbf{Taxonomy of selected key Neural Fields papers} in five major robotics application areas.
}
\vspace{-0.4cm}
\label{fig:taxonomy}
\end{figure*}
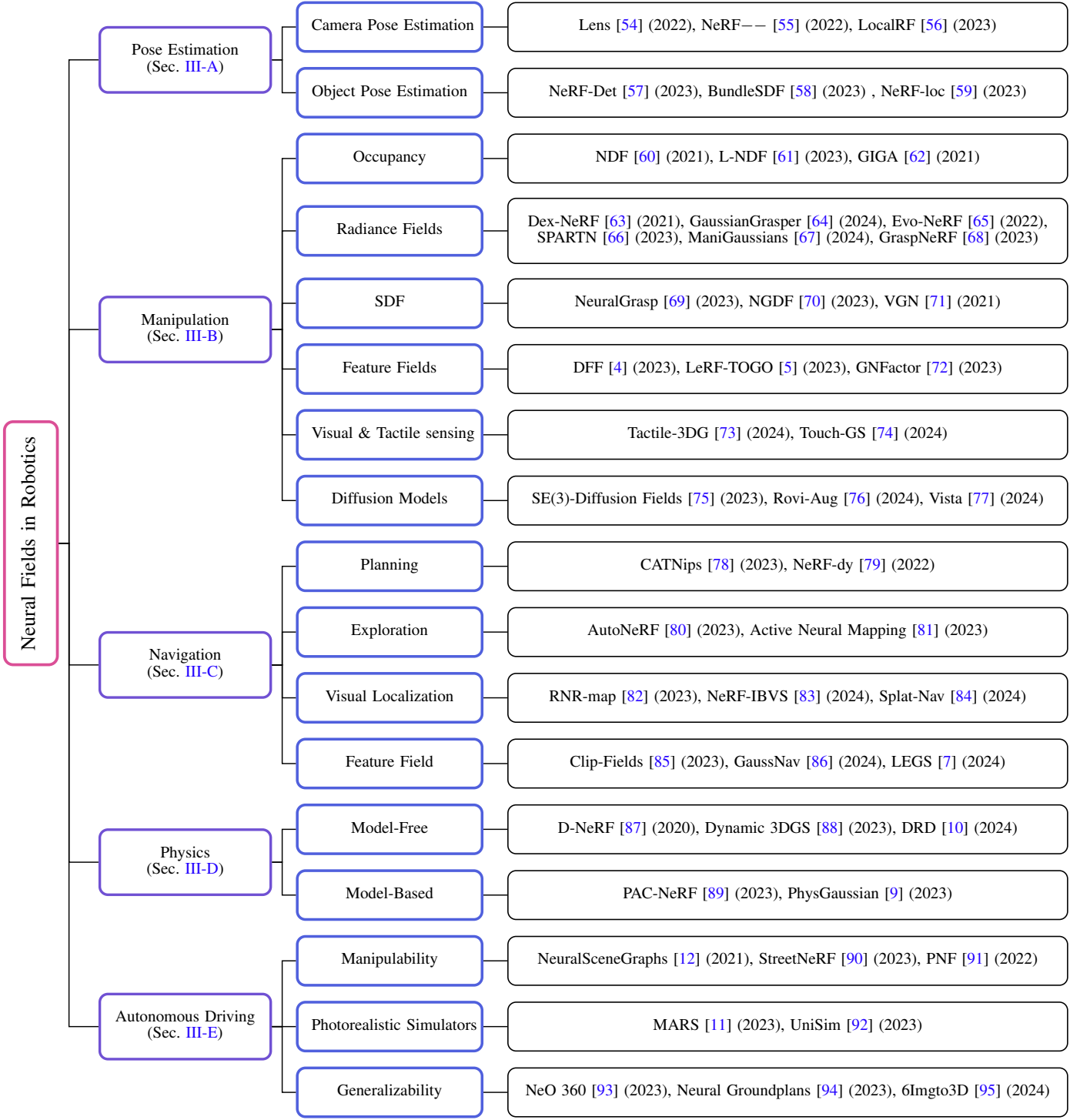

In this section, we delve into the application of Neural Fields across five major areas of robotics: pose estimation, manipulation, navigation, physics, and autonomous driving~(see \cref{fig:timeline,fig:taxonomy} for a timeline and taxonomy of selected key NFs in robotics papers). Each subsection below highlights key works within these domains, providing a comprehensive overview of the state-of-the-art methods. We conclude each subsection with a discussion of the key takeaways and the open challenges that remain in these areas, offering insights into the future directions of research in NFs for robotics.

\subsection{Neural Fields for Pose Estimation}
\label{sec:nf_pose}
Neural Fields have transformed pose estimation by offering robust and efficient methods to estimate the position and orientation of cameras and objects in 3D scenes. This section explores two key areas: camera pose estimation~(Sec.~\ref{subsec:camera_pose}) and object pose estimation~(Sec.~\ref{subsec:object_pose}). Camera pose estimation focuses on determining the viewpoint of cameras, which is crucial for tasks like mapping and reconstruction. Alternatively, object pose estimation involves localizing and orienting objects within a scene, essential for applications like manipulation and interaction. NFs optimize these tasks through gradient-based techniques, either by refining scene representations or directly providing reliable features for pose estimation.

\subsubsection{Camera Pose Estimation} \label{subsec:camera_pose}
As discussed in Sec.~\ref{Sec:Introduction}, NFs are differentiable, allowing gradient updates through scene representations, such as NeRF's volumetric rendering, down to camera parameters. While differentiable rendering has been applied to meshes~\cite{loper2014opendr, liu2019soft}, we focus here on methods applicable to NeRF-like models. This section starts by examining techniques that rely on pre-optimized NeRFs. We then explore approaches that tackle simultaneous pose estimation and geometry reconstruction, concluding with a discussion on their impact on Simultaneous Localization and Mapping (SLAM).

\looseness=-1
{\noindent \textbf{Pose Estimation via Optimized Neural Fields}: } For localization, iNeRF~\cite{yen2020inerf} inverts an already optimized NeRF for the task of pose estimation. Starting with an image, iNeRF finds the translation and rotation of a camera relative to a pretrained NeRF by using gradient descent to minimize the residual between pixels rendered from an optimized NeRF. Parallel iNeRF~\cite{lin2023icra:pnerf} parallelized the optimization processes of 6DoF poses based on fast pretrained NeRFs. Lens~\cite{moreau2022lens} prevented the generation of novel views in irrelevant areas by choosing virtual camera positions based on the NeRF’s internal 3D scene geometry. The rendered images were then used as synthetic data to efficiently train a camera pose regression model.

\looseness=-1
{\noindent \textbf{Simultaneous Pose Estimation and Reconstruction}:} 
NeRF\textminus\textminus\cite{wang2021nerf} 
and BARF~\cite{lin2021barf} show that given RGB observations of a scene, camera poses and NeRFs can be jointly optimized, removing the need for classical Structure-from-Motion~(SfM) pipelines. The former initializes cameras to the origin for forward-facing scenes, while the latter employs a coarse-to-fine reconstruction scheme that gradually introduces higher frequency position encodings~\cite{park2021nerfies, hertz2021sape}. This coarse-to-fine approach can also be adapted to multi-resolution grids like NGP~\cite{muller2022instant} by applying a weighted schedule across resolution levels~\cite{melas2023realfusion, heo2023robust}.

LocalRF~\cite{meuleman2023localrf} reconstructs long trajectories incrementally by adding images sequentially, using a subdivision approach similar to BlockNeRF~\cite{tancik2022block}, without relying on structure-from-motion (SfM). Notably, LocalRF uses different learning rates for translation and orientation parameters, highlighting the challenges in taming camera-optimizing NeRFs. GNeRF~\cite{meng2021gnerf} proposes a pose-conditioned GAN to recover a NeRF. Others have explored more suitable techniques for pose optimization, such as Gaussian~\cite{chng2022garf} or sinusoidal activations~\cite{xia2022sinerf}.
NoPe-NeRF~\cite{bian2022nope} uses monocular depth priors to constrain the scene as well as relative pose estimates. Keypoint matches or dense correspondences can also be used to constrain the relative pose estimates using ray-to-ray correspondence losses~\cite{truong2022sparf, jeong2021self}.
DBARF~\cite{chen2023dbarf} proposes using low-frequency feature maps to guide the bundle adjustment for generalizable NeRFs~\cite{wang2021ibrnet, yu2020pixelnerf, irshad2023neo360}. Nerfels~\cite{avraham2022nerfels} combines invertible neural rendering with traditional keypoint-based camera pose optimization.

\begin{figure}[t!] 
	\centering
	\includegraphics[width=1.0\columnwidth]{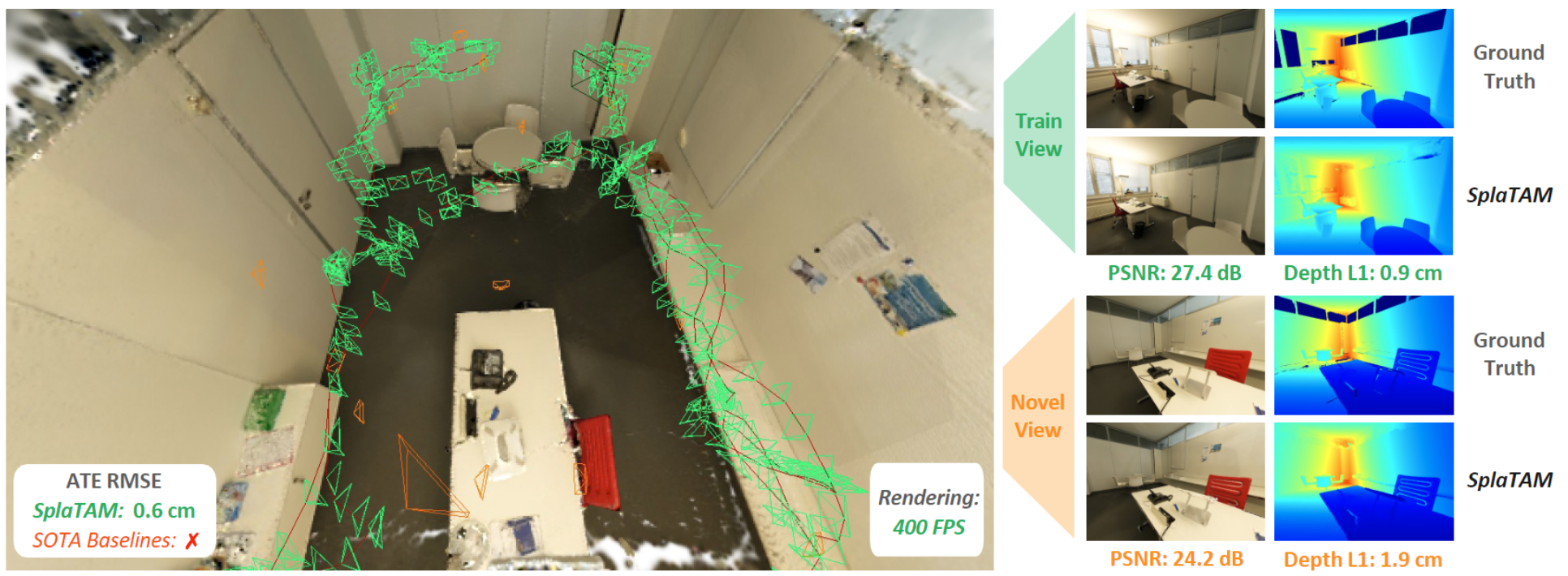}%
	\vspace{-0.5em}
	\caption{Mapping and tracking results from SplaTam~\cite{keetha2024splatam}.}
 \vspace{-0.3cm}
	\label{fig:splatam}
\end{figure}

\begin{figure}[t!] 
	\centering
	\includegraphics[width=1.0\columnwidth]{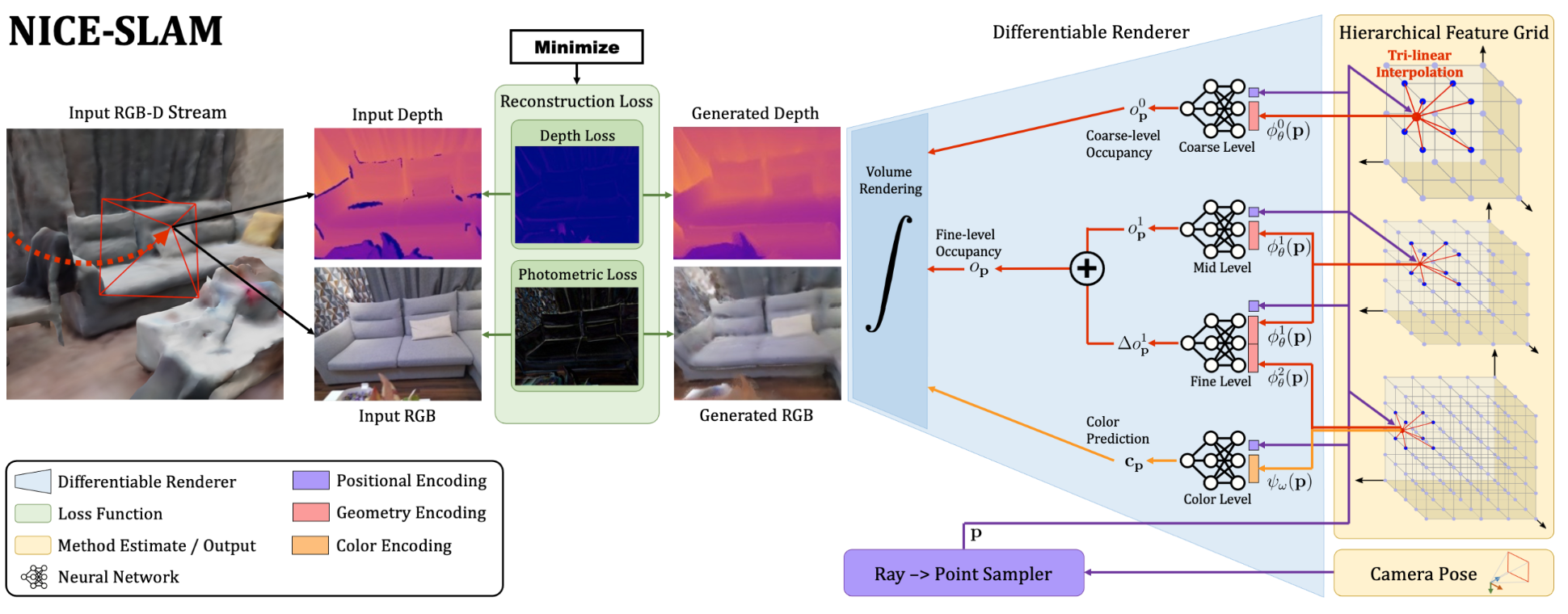}%
	\vspace{-0.5em}
	\caption{Network architecture of Nice-SLAM~\cite{zhu2022nice}.}
 \vspace{-0.5cm}
	\label{fig:niceslam}
\end{figure}

Various works that apply NFs for localization and pose estimation utilize NeRF's internal features to establish 2D-3D correspondences~\cite{zhou2024nerfect, chen2024marrying}, remove the need for an initial pose estimate~\cite{bortolon2024iffnerf}, augment the training set of the pose regressor with a few-shot NeRF~\cite{ito2023few}, or apply a decoupled representation of pose along with an edge-based sampling strategy to enhance the learning signal~\cite{claessens2023robust}. They also address dynamic scenes by integrating geometric motion and segmentation for initial pose estimation, combined with static ray sampling to speed up view synthesis~\cite{karaoglu2023dynamon}. \\
{\noindent \textbf{Simultaneous Localization and Mapping~(SLAM)}:} Jointly optimizing camera poses and neural scene representations is a fundamental aspect of the SLAM (Simultaneous Localization and Mapping) problem. Recent advancements in NFs have reshaped SLAM by leveraging their qualities, such as the ability to model continuous surfaces, lower memory usage, and enhanced robustness against noise and outliers. For instance, iMap~\cite{sucar2021imap} utilizes a single MLP to predict radiance fields as the mapping representation, employing parallel tracking and mapping threads where the tracking thread optimizes the pose of the input RGB-D frame, and the mapping thread refines both the MLP and camera pose for selected keyframes. NICE-SLAM~\cite{zhu2022nice}~(see Fig.~\ref{fig:niceslam}) further enhances this approach by replacing the single MLP with hierarchical feature grids, resulting in faster inference and more accurate reconstructions.
Gaussian splatting-based SLAM systems exploit the advantages of 3DGS including faster runtime and improved photorealistic rendering to further enhance the performance~\cite{Matsuki:Murai:etal:CVPR2024, yugay2023gaussianslam, keetha2024splatam}~(see Fig.~\ref{fig:splatam}). Besides better reconstruction quality, NF-based methods also provide an easier way to store various semantic information. Various semantic SLAM systems~\cite{Zhi:etal:arxiv2021, mazur2023feature,conceptfusion, li2024sgs} use NFs as unified representations to represent diverse information of the environment. For a more thorough survey on NFs for SLAM, we refer readers to Tosi~\etal~\cite{tosi2024nerfs}.

\subsubsection{Object Pose Estimation}\label{subsec:object_pose}
NFs have also been employed for localizing and orienting objects within a scene. Accurate pose estimation is crucial for robotics, as it enhances a robot's ability to interact with its surroundings and perform tasks such as manipulation, navigation, and object recognition. Works in this domain use NFs' features to establish correspondences or directly regress poses and reconstruct shapes. This facilitates the determination of bounding boxes and orientations for various objects in a 3D environment. 

\begin{figure}[t!] 
	\centering
	\includegraphics[width=1.0\columnwidth]{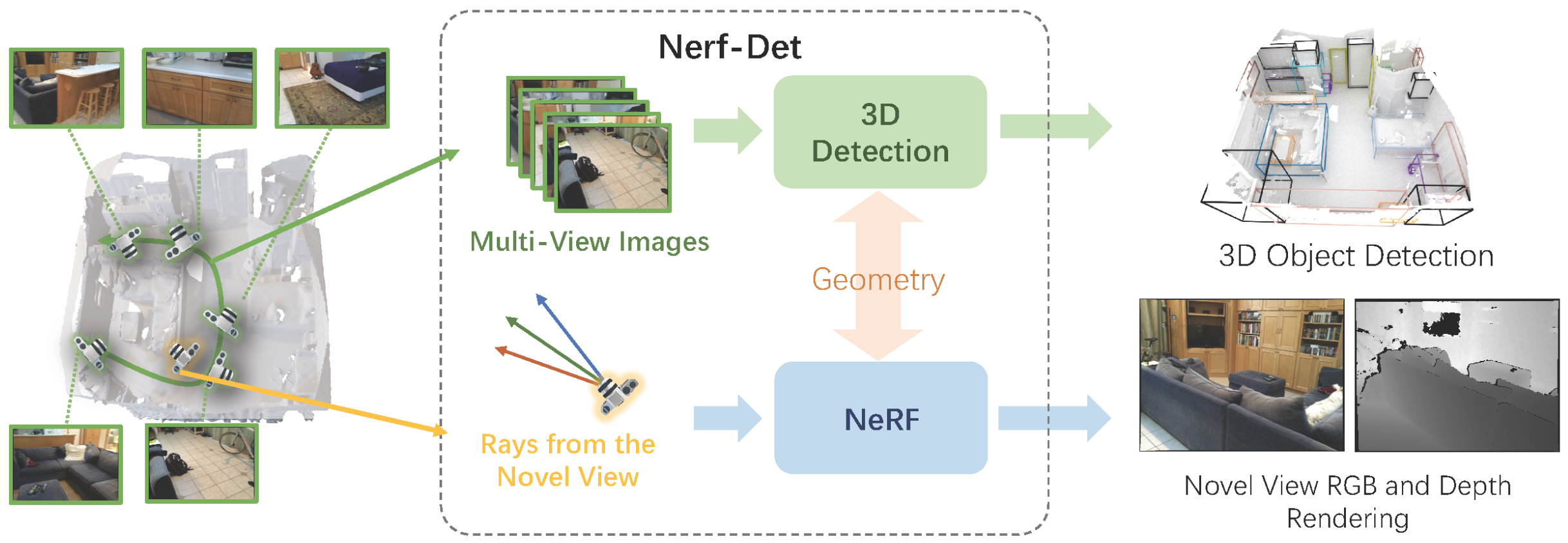}%
	\vspace{-0.5em}
	\caption{NeRF-Det's~\cite{xu2023nerfdet} 3D detection pipeline using NeRFs.}
 \vspace{-1.5em}
	\label{fig:nerfdet}
\end{figure}

\looseness=-1
Neural Field features are also shown to be effective for multi-view 3D bounding box estimation of objects in the scene. NeRF-RPN~\cite{hu2023nerf} estimates 3D object boxes directly on NeRF's feature grid, using a novel voxel representation and without re-rendering from a pretrained NeRF. It can be trained end-to-end to estimate high-quality 3D bounding boxes without class labels. Similarly, NeRF-Det~\cite{xu2023nerfdet}~(see Fig.~\ref{fig:nerfdet}) leverages a NeRF to explicitly estimate 3D geometry and improve detection performance. It introduces geometry priors and connects detection with NeRF branches through a shared MLP, enhancing generalizability and efficiency without per-scene optimization. NeRF-RPN's performance can be further enhanced with self-supervised representation learning directly using NeRF grids, as shown by NeRF-MAE~\cite{irshad2024nerfmae}. Similarly, Gaussian splats~\cite{kerbl20233d} have also been effectively utilized for 3D object detection, as demonstrated by GaussianDet~\cite{yan2024gaussiandetlearningclosedsurfacegaussians} and 3D-GSDet~\cite{cao20243dgsdetempower3dgaussian}. 
Furthermore, NeRF-loc~\cite{sun2023nerf} employs a transformer-based framework to extract labeled, oriented 3D bounding boxes of objects from NeRF scenes. It utilizes a pair of parallel transformer encoder branches to encode both the context and details of target objects, fusing these features with attention layers for accurate object localization, outperforming conventional RGB(-D) based methods. NeRF-pose~\cite{li2023nerf} employs a weakly supervised 6D pose estimation pipeline that requires only 2D segmentation and known relative camera poses during training, avoiding the need for precise 6D pose annotations. It reconstructs objects from multiple views, then trains a pose regression network to predict 2D-3D correspondences with a NeRF-enabled Perspective-n-Point~(PnP)+RANSAC algorithm estimating stable and accurate poses from a single input image.

\begin{figure}[t!] 
	\centering
\includegraphics[width=1.0\columnwidth]{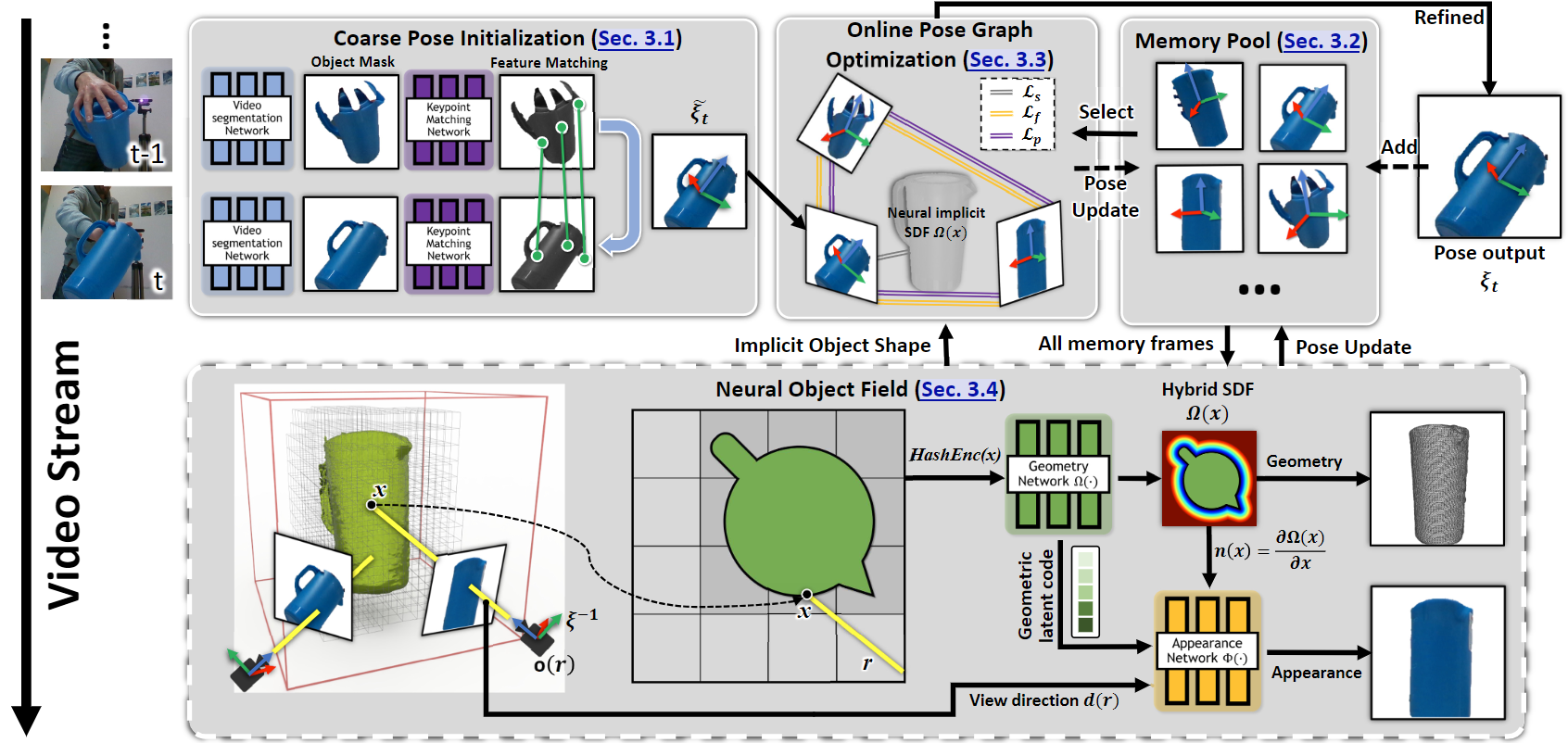}%
	\vspace{-0.5em}
	\caption{BundleSDF~\cite{wen2023bundlesdf} for object tracking and reconstruction.}
 \vspace{-0.4cm}
	\label{fig:bundlesdf}
\end{figure}

Other direct pose-regression methods also simultaneously reconstruct object shapes in conjunction with object pose estimation. ShAPO~\cite{irshad2022shapo}, FSD~\cite{lunayach2023fsd} and CARTO~\cite{heppert2023carto} jointly reconstruct object shapes and regress their 6D object poses using a single-shot pipeline employing implicit representations and disentangled shape and appearance priors. UPNeRF~\cite{guo2024upnerf} proposes a unified framework for monocular 3D reconstruction that integrates pose estimation with NeRF-based reconstruction, addressing the shortcomings of existing methods that rely on external 3D object detectors for initial poses. It decouples dimension estimation and pose refinement to resolve scale-depth ambiguity, and employs a projected-box representation for cross-domain generalization. NeRF-from-image~\cite{pavllo2023shape} integrates NeRF with GANs to model arbitrary topologies without requiring accurate ground-truth poses or multiple views during training. It uses an unconditional 3D-aware generator and a hybrid inversion scheme to recover an SDF-parameterized 3D shape, pose, and appearance, refining initial solutions via optimization. NCF~\cite{huang2022neural} estimates the 6DoF pose of a rigid object with a 3D model from a single RGB image by predicting 3D object coordinates at 3D query points sampled in the camera frustum rather than at image pixels. Bundle-SDF~\cite{wen2023bundlesdf}~(see Fig.~\ref{fig:bundlesdf}) tracks the 6-DoF pose of an unknown object from a monocular RGBD video sequence while simultaneously performing neural 3D reconstruction. It handles arbitrary rigid objects with minimal visual texture, requiring only the object's segmentation in the first frame. The approach uses a Neural Object Field learned alongside pose graph optimization to build a consistent 3D representation of the object's geometry and appearance.

\subsubsection{Takeaways and Open Challenges in Neural Fields for Pose Estimation}
\label{key_takeaway_pose}

Despite the promising progress in NFs for pose estimation, several open challenges remain. Current methods prove effective in real-time pose estimation of cameras as well as objects. While significant progress has been made in pose estimation for static scenes, there is still room for further exploration in dynamic environments. Future work could focus on refining methods for recovering camera poses from dynamic video capture, where both cameras and objects exhibit significant movement, such as post-hoc calibration and labeling of robotic datasets~\cite{khazatsky2024droid}. This would open up the possibility of learning 3D priors from large-scale monocular videos. Another avenue for future work could explore using NFs for open-vocabulary 6D object pose estimation and solving the canonicalization problem of large-scale datasets~\cite{deitke2024objaverse}. 

\subsection{Neural Fields for Robotic Manipulation}
\label{sec:nf_manipulation}

One of the key challenges in robotic manipulation is obtaining a precise geometric representation of both the objects and the environment involved in the task. An effective representation must also capture the environment dynamics, offering a robust 3D understanding of the objects. In this section, we explore the application of NFs in control tasks for manipulation, with a focus on pick-and-place scenarios within 3 and 6 Degrees of Freedom (DoF). A summary of methods leveraging NFs for manipulation is provided in Table~\ref{tab:papers_grasping}.

Approaches that synthesize 3-DoF grasps use visual input, like RGB or depth images, to generate grasps in the image frame~\cite{breyer2021volumetric}. This means the end effector can translate horizontally and rotate around its vertical axis, with depth sensors determining vertical positioning. However, 3-DoF grasping lacks precise orientation control. In contrast, 6-DoF methods predict both position and orientation using 3D representations, enabling full control over roll, pitch, and yaw, which enhances the robot's ability to manipulate objects in any direction.

\begin{figure}[t!] 
	\centering
	\includegraphics[width=1.0\columnwidth]{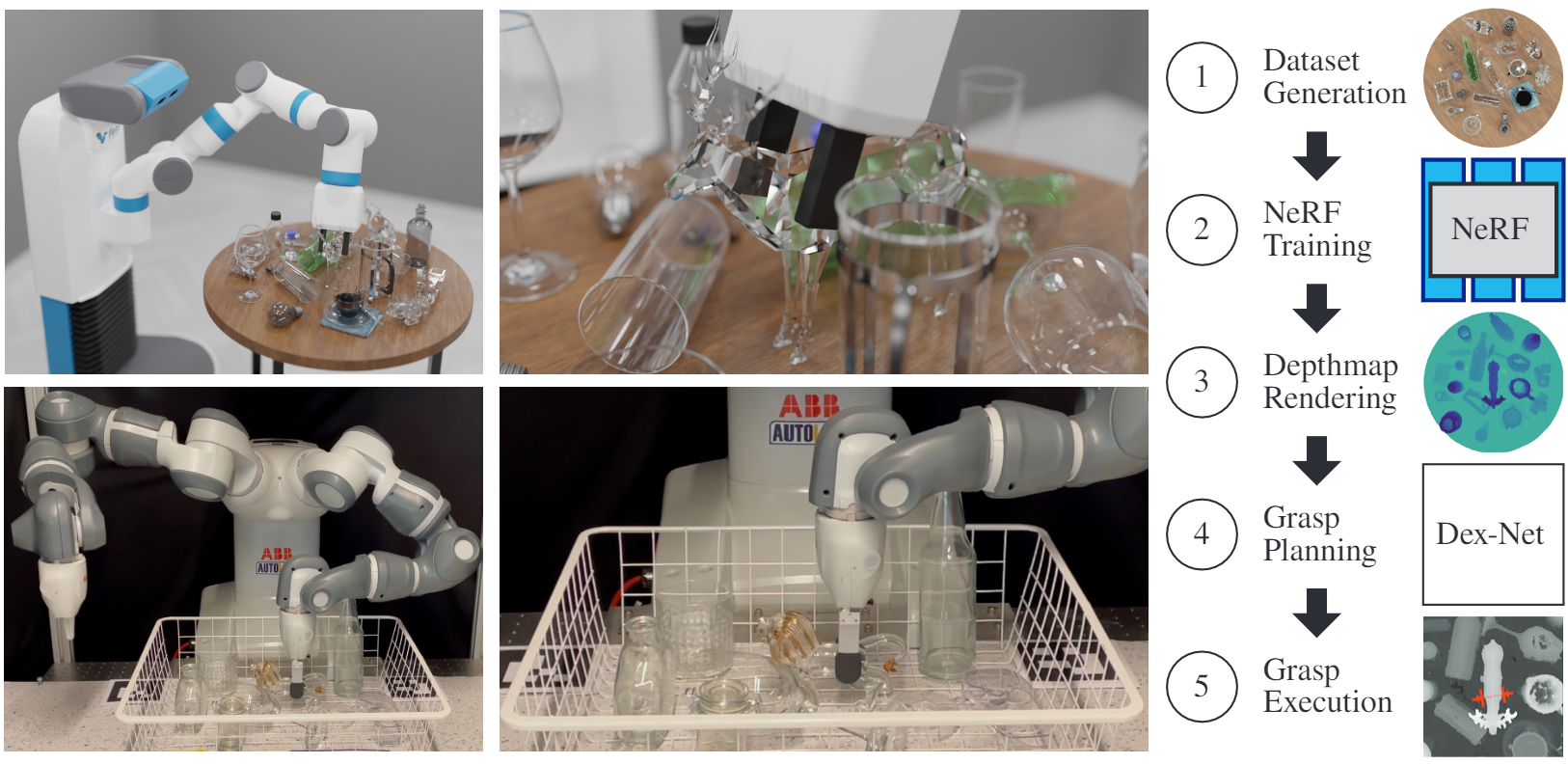}%
	\vspace{-0.5em}
	\caption{Dex-NeRF~\cite{ichnowski2021dex} leverages NeRF's depthmap rendering for transparent object grasping.}
 \vspace{-0.4cm}
	\label{fig:dexnerf}
\end{figure}

Within the scope of 3-DoF, Dex-NeRF~\cite{ichnowski2021dex}~(see Fig.~\ref{fig:dexnerf}) uses NFs to detect and infer the geometry of transparent objects, employing a transparency-aware rendering technique and additional lighting for specular reflections. Combined with Dex-Net~\cite{mahler2017dex}, it generates 3-DoF grasping poses for transparent objects in both simulated and real environments. More recently, Evo-NeRF~\cite{kerr2022evo} extends this by leveraging Instant-NGP~\cite{muller2022instant} to accelerate inference and adapt NeRF weights for sequential grasping tasks on transparent objects, updating the representation with each grasp.

Recently, there has been a shift toward using NFs for 6-DoF grasp pose estimation, offering an alternative to traditional point cloud-based methods. Below, we discuss these representations in detail:

\begin{table*}[h]
    \small
    \centering
    \begin{tabular}{c c c c c}
    \toprule
    \multirow{2}{*}{\textbf{Fields}} & \multirow{2}{*}{\textbf{Input}} & \multirow{2}{*}{\textbf{Method}} & \multirow{2}{*}{\textbf{Representation}} & \multirow{2}{*}{\textbf{Scope}} \\
     & & & & \\

    \midrule
    \multirow{2}{*}{Occupancy} & \multirow{2}{*}{Point-cloud}  & NDF~\cite{simeonov2022neural} &  VNN, ONet & Scene-specific \\
    & & L-NDF~\cite{chun2023local} &  VNN, ONet & General \\

    \midrule

    \multirow{5}{*}{Signed Distance} & Single-view Depth & GIGA~\cite{jiang2021synergies} &  ConvONet, TSDF &  General \\
    \cmidrule{2-5}

    & \multirow{2}{*}{Multi-view Depth} & VGN~\cite{breyer2021volumetric} & TSDF & General \\
    & & NGDF~\cite{weng2023neural} & SDF & General \\
    \cmidrule{2-5}

   & \multirow{2}{*}{Multi-view RGB-D}& Song \etal~\cite{song2020grasping} & TSDF & General\\
    & & NeuralGrasps~\cite{khargonkar2023neuralgrasps} & SDF & Object-specific \\
    \midrule

    \multirow{12}{*}{Radiance} & \multirow{8}{*}{Multi-view RGB} &  Dex-NeRF~\cite{ichnowski2021dex} & NeRF & Scene-specific \\
    & & Evo-NeRF~\cite{kerr2022evo} & Instant-NGP &  Scene-specific \\
    & & NeRF-Supervision~\cite{yen2022nerf} &  NeRF  & Scene-specific \\
    & & MIRA~\cite{yen2022mira} & NeRF, Orthographic images & Scene-specific \\
    & & SPARTN~\cite{zhou2023nerf} & NeRF & Scene-specific \\ 
    & & MVNeRF~\cite{soti20246} & NeRF & General \\
    & & RGBGrasp~\cite{liu2023rgbgrasp} & NeRF, Hash encoding & General \\
    & & GraspNeRF~\cite{dai2023graspnerf} & Generalizable NeRF, TSDF & General \\
    \cmidrule{2-5}
    & \multirow{2}{*}{Multi-view RGB-D} &
    GaussianGrasper~\cite{zheng2024gaussiangrasper} & 3DGS & General \\
    
    & & ManiGaussian~\cite{lu2024manigaussian} & 3DGS & General \\
    \cmidrule{2-5}
    
    & Single-view RGB, Annotations & Blukis \textit{et al.}~\cite{blukis2022neural, blukis2023one} &  NeRF  & General \\
    \bottomrule
    \end{tabular}
    \caption{Overview of selected methods that leverage neural fields for manipulation tasks. See Sec.~\ref{sec:nf_manipulation} for more details.}
    \vspace{-0.4cm}
    \label{tab:papers_grasping}
\end{table*}

\subsubsection{Occupancy Fields}

Neural Descriptor Fields (NDF)~\cite{simeonov2022neural} propose an SE(3)-equivariant object representation for manipulating novel objects in arbitrary poses, with few demonstrations. Using a Vector Neurons (VN)~\cite{deng2021vector} network, 6-DoF relative poses between objects and grippers are encoded. NDF represents objects as continuous 3D descriptor fields, mapping points to descriptor vectors that capture spatial relationships to object geometry. However, NDF struggles with generalizing to new object categories, a limitation addressed by Local Neural Descriptor Fields (L-NDF)~\cite{chun2023local}, which use a voxel grid of local embeddings to better capture local geometry and descriptors for new shapes. Both NDF and L-NDF rely on VN equipped with an Occupancy Network (ONet). GIGA~\cite{jiang2021synergies} leverages a Convolutional Occupancy Network (ConvONet) to detect 6-DoF grasps in cluttered environments from a single depth image. By encoding the Truncated Signed Distance Function (TSDF), GIGA jointly predicts volumetric occupancy and 6-DoF grasp detection, enabling it to detect grasps on occluded objects from partial observations.

\subsubsection{Radiance Fields} These fields are primarily modeled using two approaches: NeRFs and 3D Gaussian Splatting.

\noindent\textbf{NeRF:} 
NeRF-Supervision~\cite{yen2022nerf} leverages NeRFs to generate synthetic data for dense correspondence estimation by treating correspondences as depth distributions instead of pixel-wise depth. This enables 6-DoF picking tasks on thin and reflective surfaces using only RGB images, though multi-view images are required to build the NeRF representation. MIRA~\cite{yen2022mira} extends this by constructing a NeRF before each action, enabling pick-conditioned placing via view synthesis. MIRA also trains a NeRF model with perspective cameras for direct orthographic rendering, which aligns well with translationally equivariant architectures like ConvNets. SPARTN~\cite{zhou2023nerf} enhances visual grasping policies by using synthetic multi-view RGB images from eye-in-hand camera setups, significantly improving success rates in grasping tasks over standard imitation learning methods. These advancements highlight NeRF's potential to bridge the gap between RGB-based and depth-based robotic policies.

Blukis~\etal~\cite{blukis2022neural, blukis2023one} proposed an approach that jointly optimizes 3D reconstruction and grasp pose estimation by encoding objects into a unified latent representation. Encoded latents are decoded for view synthesis, 3D reconstruction, and grasp proposals. NeRFs have also been adapted for transfer learning in 6-DoF grasp pose evaluation and optimization with MVNeRF~\cite{soti20246}, which processes inputs from multiple scenes, enabling a more generalized representation and faster perception-to-action mapping. RGBGrasp~\cite{liu2023rgbgrasp} advances real-time applications by integrating multi-view RGB data from an eye-on-hand camera and depth maps from a pre-trained model. It further accelerates 3D reconstruction using hash-encoding~\cite{muller2022instant} and a novel sampling strategy. 

\noindent\textbf{3D Gaussian Splatting:} %
\label{subsec:manip:3D_gs}%
The introduction of 3D Gaussian Splatting~(\cref{sec:3dgs}) marks a promising advancement in leveraging 3D representations for real-time robotic manipulation. GaussianGrasper~\cite{zheng2024gaussiangrasper} proposes a novel approach to 6-DoF grasping using Gaussian Splatting for open-vocabulary object grasping, thus enabling robots to understand and execute tasks based on natural language instructions. Similarly, ManiGaussian~\cite{lu2024manigaussian} builds on dynamic 3D Gaussian Splatting to capture scene-level spatiotemporal dynamics, enhancing the robot's capability to execute tasks conditioned on natural language instructions. SplatSim~\cite{qureshi2024splatsim} shows an application of 3DGS for improving Sim2Real transfer for robotic manipulation policies that rely on RGB images. This is obtained by leveraging the photorealism of 3DGS, which reduces the domain shift between synthetic and real visual information.

\subsubsection{Signed Distance Fields}
\begin{figure}[t!] 
	\centering
	\includegraphics[width=1.0\columnwidth]{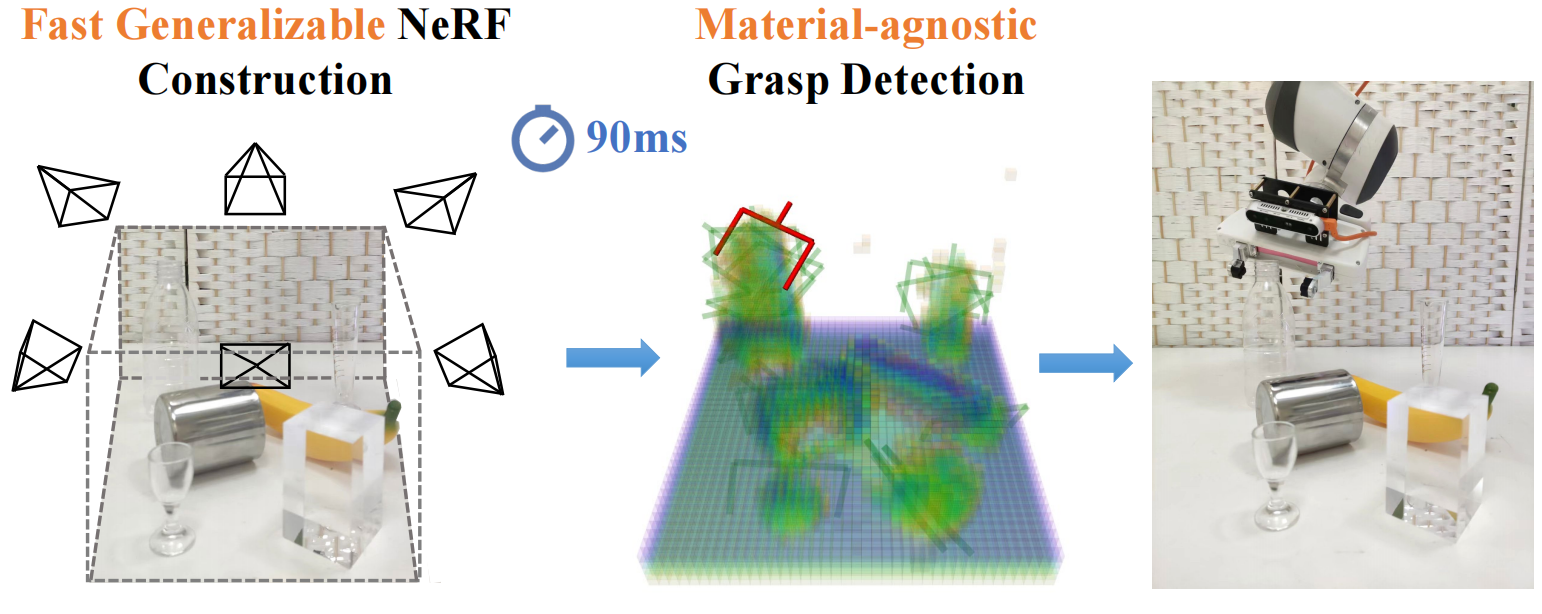}%
	\vspace{-0.5em}
	\caption{Generalizable grasping with sparse multi-view images using GraspNeRF~\cite{dai2023graspnerf}.}
 \vspace{-1em}
	\label{fig:graspnerf}
\end{figure}

GraspNeRF~\cite{dai2023graspnerf}~(see Fig.~\ref{fig:graspnerf}) extends NeRF for 6-DoF grasp detection of transparent and specular objects. It integrates a Truncated Signed Distance Function (TSDF) with a generalizable NeRF trained on sparse RGB images for zero-shot scene reconstruction. Similarly, Volumetric Grasping Network (VGN)~\cite{breyer2021volumetric} enables real-time 6-DoF grasp detection, synthesizing collision-free grasps in cluttered environments using a 3D voxel grid representation where each voxel contains the TSDF to the nearest surface. Song \etal~\cite{song2020grasping} employ a TSDF to map actions to rendered views, simulating future state-action pairs for 6-DoF grasping. NeuralGrasps~\cite{khargonkar2023neuralgrasps} further explores neural distance fields by learning implicit representations for grasps with multiple robotic hands. This method encodes grasps into a shared latent space, with each vector corresponding to a grasp from a specific hand. The neural Grasp Distance Field (NGDF)~\cite{weng2023neural} models grasp poses as the level set of an unsigned distance field, predicting the closest valid grasp for a given 6D query pose by minimizing the unsigned distance. CenterGrasp~\cite{chisari2024centergrasp} extends this concept to directly predict a displacement vector, removing the need to optimize the level-set.

\subsubsection{Feature Fields}
Feature fields represent an emerging class of NFs that integrate high-dimensional features from visual data into a unified 3D representation. These fields can map 3D points to feature vectors encoding semantic information, making them useful for context-aware grasping tasks when combined with pre-trained vision-language models.

\begin{figure}[t!] 
	\centering
	\includegraphics[width=1.0\columnwidth]{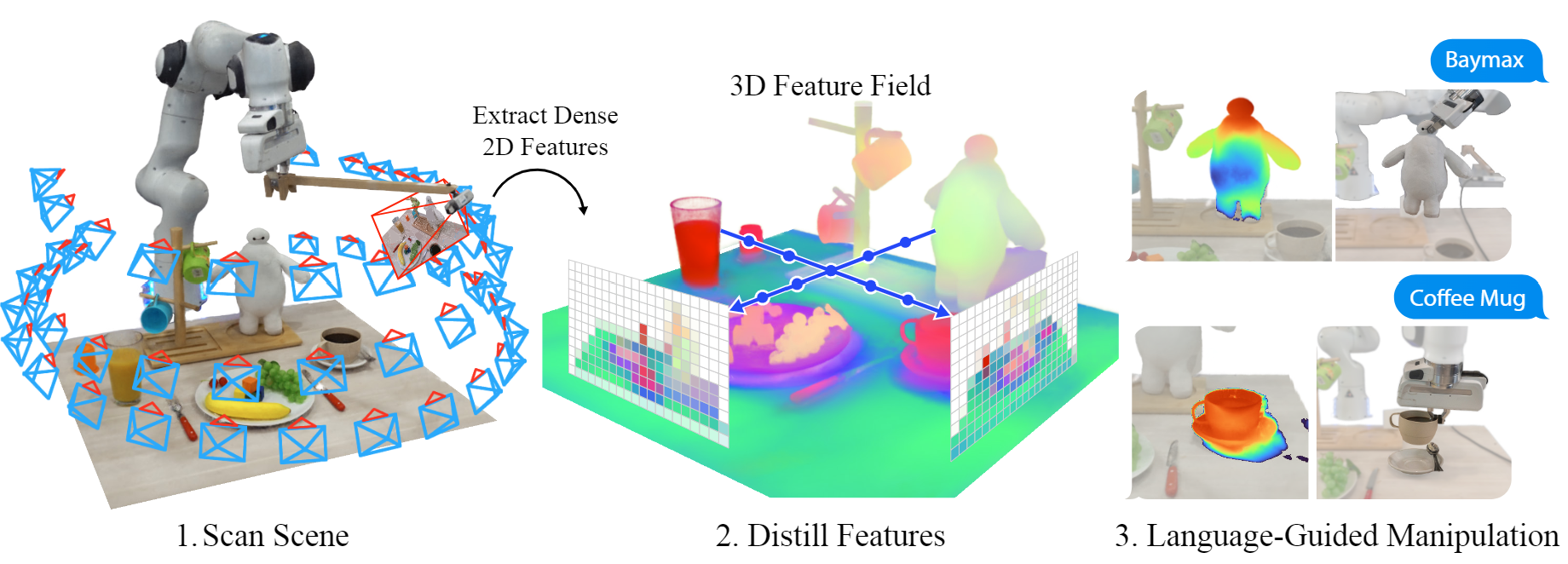}%
	\vspace{-0.5em}
	\caption{Distilled feature fields~\cite{shen2023F3RM} distill foundation model features into a feature field along with modeling a NeRF.}
 \vspace{-0.4cm}
	\label{fig:f3rm}
\end{figure}

Developments in this area have focused on the creation of feature fields that enable few-shot learning and zero-shot task-oriented grasping. \textit{Distilled Feature Fields} (DFF)~\cite{shen2023F3RM} proposes to distill dense features from a pre-trained vision-language model (CLIP~\cite{radford2021learning}) into a 3D feature field (see~\cref{fig:f3rm}). This allows for effective generalization across diverse object categories, making DFF particularly useful for tasks that require contextual understanding. Similarly, LERF-TOGO~\cite{lerftogo2023} leverages language embeddings and DINO~\cite{oquab2023dinov2} features to accurately select target objects and specific object parts for grasping. This intuition addresses the limitations of traditional learning-based grasp planners that often ignore the semantic properties of objects. The use of vision-language models for feature distillation is also proposed by GeFF~\cite{qiu2024learning} and GNFactor~\cite{ze2023gnfactor}, where a generalizable NeRF enriched by semantic information is adopted for manipulation and navigation. In conclusion, by incorporating semantic information directly into the 3D representation, feature fields facilitate precise, context-aware, language-guided manipulation in real-world scenarios.

\subsubsection{Visual \& Tactile Sensing}
The use of NFs in multimodal visual and tactile sensing is a recent and emerging area of research. Tactile data gathered from tactile sensors offers information about contact force and contact geometry. Combining visual and tactile sensing offers several advantages in situations where vision-only might be ambiguous, such as in the presence of occlusions, challenging lighting, or reflective and transparent materials~\cite{comi2024snap, swann2024touch}. In the context of multimodal sensing, NFs are mainly leveraged for tactile data generation and object reconstruction. Zhong~\etal~\cite{zhong2023touching} propose to use NeRF to generate realistic tactile sensory data. Similarly, TaRF~\cite{dou2024tactile} leverages NeRF to synthesize novel views, which are subsequently used by a conditional diffusion model to generate the corresponding tactile signal. 

A challenge in using tactile data is the disparity between real and simulated tactile images. TouchSDF~\cite{comi2024touchsdf} addresses this sim-to-real gap by combining a Convolutional Neural Network (CNN) with DeepSDF for 3D shape reconstruction from tactile inputs. As a result, objects can be reconstructed using solely tactile sensing in both simulation and the real world. Suresh \etal~\cite{suresh2023neural} employ a neural signed distance field to estimate object pose and shape during in-hand manipulation. The use of NFs, in this context, allows the robot to learn and progressively refine the object's shape online. Moreover, NFs and visuo-tactile sensing can be employed for predicting extrinsic forces. Neural Contact Fields (NCF)~\cite{higuera2023neural} tracks the contact points between an object and its environment using tactile information. This method leverages NFs to generalize across different object shapes and to estimate the probability of contact at any point on an object’s surface. 

Although most studies focus on implicit representations, recent techniques have begun exploring explicit approaches. Tactile-3DGS~\cite{comi2024snap} and Touch-GS~\cite{swann2024touch} extend 3D Gaussian Splatting by integrating visual and tactile data for 3D object reconstruction. Unlike TouchSDF, both Touch-GS and Tactile-3DGS are designed to handle the reconstruction of transparent and reflective objects. These methods have been validated in both simulated and real-world environments.

\subsubsection{Diffusion Models}
Recent work in Generative AI for robotics manipulation has explored the use of diffusion models for grasp generation, trajectory planning, and learning view-point and cross-embodiment invariant policies. Yoneda \textit{et al.}~\cite{yoneda20236} leverage diffusion models to predict stable object placements by learning context-dependent distributions from positive examples, eliminating the need for rejection sampling. SE(3)-Diffusion Fields~\cite{urain2023se3diffusionfieldslearningsmoothcost} optimize grasp selection and trajectory generation by learning data-driven SE(3) cost functions using diffusion models. Meanwhile, VISTA~\cite{tian2024vista} and RoVi-Aug~\cite{chen2024roviaug} utilize 3D generative models for learning viewpoint-invariant policies, enabling robust generalization to new environments and unseen robots. VISTA~\cite{tian2024vista} leverages Zero-NVS~\cite{zeronvs}'s zero-shot novel view synthesis capability to learn viewpoint-invariant policies, enabling robust performance in diverse environments and tasks from limited demonstration data. Similarly, RoVi-Aug~\cite{chen2024roviaug} synthesizes augmented robot data using image-to-image generative models, allowing for zero-shot deployment on unseen robots with different embodied and largely varying camera angles. Together, these methods illustrate how 3D generative techniques can significantly improve the adaptability and effectiveness of robotic manipulation systems in real-world scenarios.

\subsubsection{Takeaways and Open Challenges in Neural Fields for Manipulation}
NFs have emerged as powerful techniques for robust 3D understanding in robotic manipulation tasks, such as grasping and pick-and-place. These representations capture detailed geometrical information and support generalization across diverse object shapes and categories. NFs have also been employed to identify optimal grasp points, improving the success rate of robotic grasps in cluttered environments. Additionally, some methods integrate these representations with language models, enabling open-vocabulary manipulation through natural language instructions.

Despite these advancements, significant challenges remain. Current approaches rely on extensive multi-view inputs or costly per-scene optimization, limiting their applicability in complex, dynamic, or unstructured environments. Furthermore, incorporating physical intuitions about object affordances and robot dynamics into the learned representations could lead to more physically grounded manipulation policies (see~\cref{sec:nf_physics}). Finally, scaling these methods to dynamic scenes with multiple agents or articulated objects is an ongoing challenge that must be addressed for real-world deployment.

\subsection{Neural Fields for Navigation}
\label{sec:nf_navigation}

Autonomous navigation requires robots to perceive and model their surroundings effectively to plan collision-free paths. Traditional learning-based approaches tackle this challenge via either end-to-end~\cite{ma2019theregretful, wang2019reinforced, wijmans2020ddppo} or modular systems~\cite{irshad2021hierarchical, irshad2022sasra, chaplot2020semantic}. Recently, the properties of NFs have proven beneficial for motion planning and navigation. NeRF’s density grid, for instance, offers a geometric approximation of the scene that aids in avoiding obstacles or learning a dynamics model. Various NeRF extensions have been proposed for navigation; some construct maps representing the visual structure of the scene, and others use autonomous agents to actively map the environment. Below, we highlight these state-of-the-art advances, structured into four key areas: Planning, Exploration, Visual Localization, and Feature Fields.

\subsubsection{Planning} 
\looseness=-1
NFs' density grid provides an approximate geometry, which is then used with a trajectory planner and state estimator in an iterative receding horizon loop for an autonomous agent to dynamically maneuver an environment with RGB camera for feedback~\cite{adamkiewicz2022vision}. CATNIPS~\cite{chen2023catnips} enables collision avoidance in a NeRF by computing collision probabilities for a robot navigating through a NeRF. It enables fast trajectory optimization using graph-based searching with spline-based trajectory optimization. SAFER-Splat~\cite{chen2024safer} provides a real-time reconstruction method using Gaussian Splatting for safe robotic navigation. It operates efficiently with minimal memory, ensuring safety while maintaining high-speed performance during online mapping. NeRF's 3D scene representation also allows learning 3D dynamical models purely from posed 2D images. Specifically, NeRF-dy~\cite{li20223d} proposed a time-contrastive learning objective, which, when combined with Neural Radiance Fields in an auto-encoding framework, provides a viewpoint-aware neural 3D scene representation. This scene representation allows for the specification of goal points with learned predictive dynamical forward models outside the training-distribution viewpoints. CompNeRFdyn~\cite{driess2023learning} extends the concepts proposed by \mbox{Li \textit{et al.}~\cite{li20223d}}, introducing an auto-encoder framework in conjunction with a Graph Neural Network~(GNN)~\cite{wu2020comprehensive} based dynamic model prediction in the latent space of NeRF. This forces the network to learn generalizable priors, thus aiding long-range predictions. 

\subsubsection{Exploration} 
\begin{figure}[t!] 
	\centering
	\includegraphics[width=1.0\columnwidth]{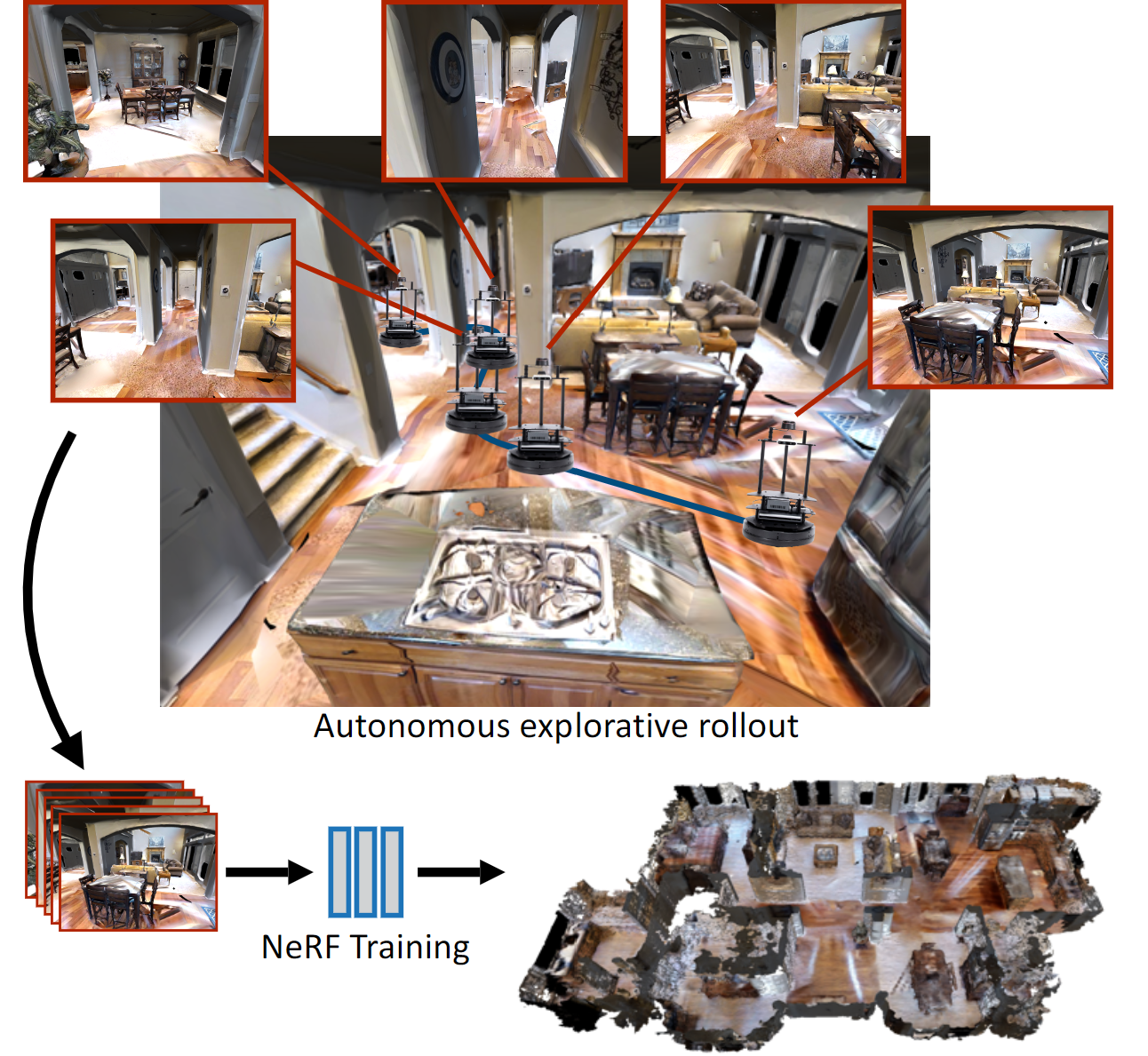}%
	
	\caption{AutoNeRF~\cite{marza2023autonerf} generates 3D models of a scene by training NeRFs from data collected by
autonomous agents.}
 \vspace{-1.5em}
	\label{fig:autonerf}
\end{figure}

\begin{figure}[t!] 
	\centering
	\includegraphics[width=1.0\columnwidth]{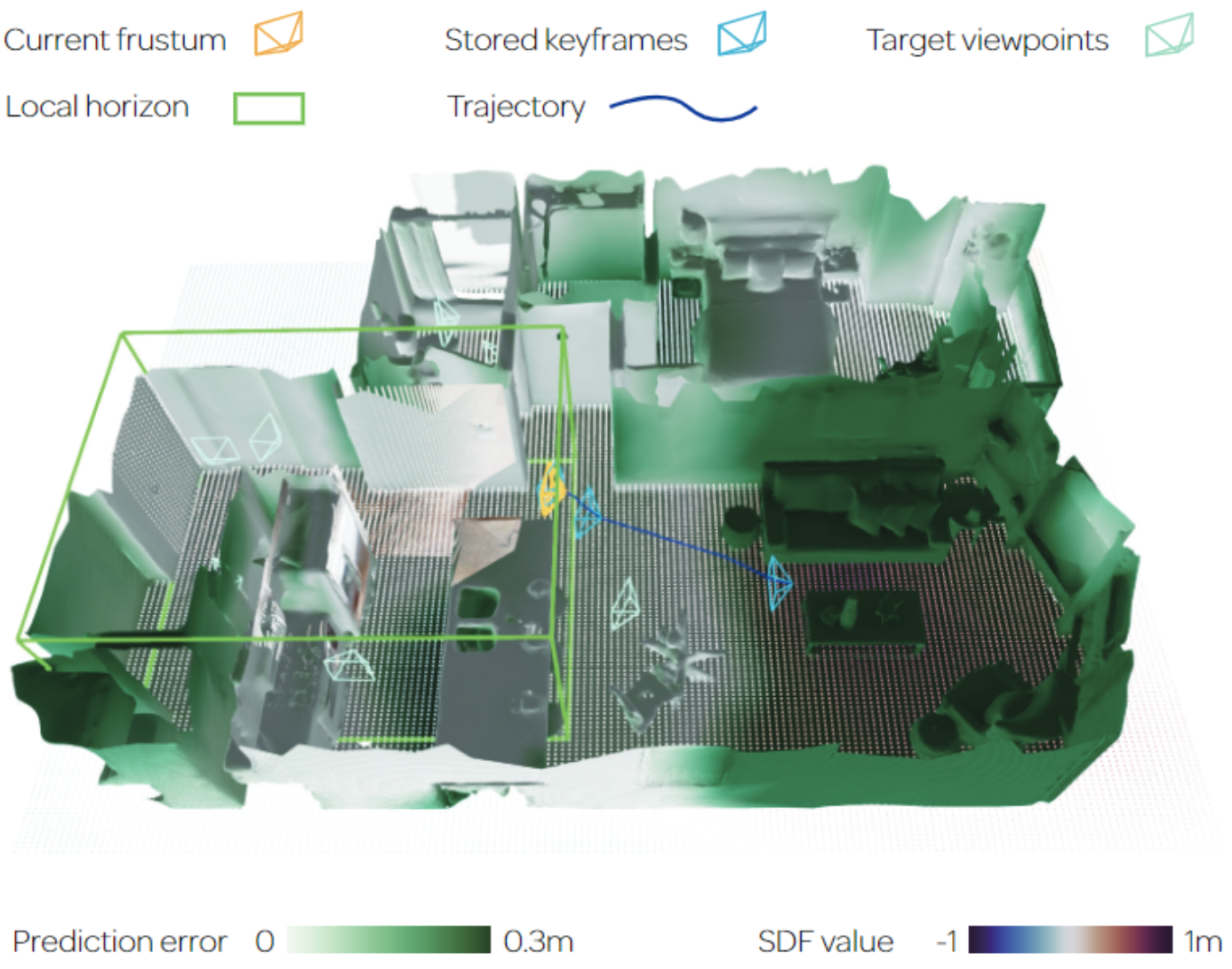}%
	\vspace{-0.5em}
	\caption{Active exploration of a mobile robot to minimize prediction uncertainty~\cite{Yan2023iccv}.}
 \vspace{-0.5cm}
	\label{fig:active_neural_mapping}
\end{figure}

A separate line of work uses modular autonomous navigation agents~\cite{chaplot2020object} to enable training of implicit scene representations~\cite{marza2023autonerf}. AutoNeRF~\cite{marza2023autonerf}~(see Fig.~\ref{fig:autonerf}) forgoes the need for carefully curated manual dataset creation with autonomously created dataset. An autonomous agent explores an unseen environment without prior access to a map and uses the experience to create implicit scene representations for novel view and semantic synthesis. DroNeRF~\cite{patel2023dronerf} enables active reconstruction and proposed a novel optimization approach to create automated positioning of cameras for few-view reconstruction of objects in an implicit manner. Active Neural Mapping~\cite{Yan2023iccv}~(see \cref{fig:active_neural_mapping}) studies the problem of actively exploring an environment with a continually learned 3D scene representation, such as a NeRF. It minimizes the map uncertainty in real-time by actively selecting target spaces for exploration. By utilizing continuous geometric information encoded in the neural map, the system guides agents to find traversable paths for online scene reconstruction. DISORF~\cite{li2024disorf} introduces a framework designed to facilitate real-time 3D reconstruction and visualization of scenes captured by resource-constrained mobile robots and edge devices. It addresses compute limitations and network constraints by distributing computation efficiently between the edge device and a remote server. The framework utilizes on-device SLAM systems to generate posed key frames, which are then transmitted to remote servers for high-quality 3D reconstruction and visualization using NeRF models. Finding Waldo~\cite{skartados2024finding} proposes baseline methods, Guided-Random Search (GRS) and Pose Interpolation-based Search (PIBS), and formulates scene exploration as an optimization problem, presenting Evolution-Guided Pose Search (EGPS) as an efficient solution.

\subsubsection{Visual Localization}
Building on the camera localization methods discussed in Sec.~\ref{sec:nf_pose}, other works utilize implicit neural representations for top-down memory, real-time navigation, and visual localization. These approaches demonstrate the application of NFs for visual localization, which is crucial for effective navigation in dynamic environments.

RNR-Map~\cite{Kwon_2023_CVPR} constructs a visually descriptive map of the environment, similar to Incremental Scene Synthesis~\cite{planche2019incremental}, where a latent code at each pixel in the grid cell is embedded from an image observation and can be converted to a Neural Radiance Field. This Radiance Field can be rendered with arbitrary camera poses. A modular framework utilizing the visual information in these maps enables visual localization as well as navigation. The Le-RNRMap model~\cite{Taioli_2023_ICCV} improves upon RNR-Map by integrating CLIP-based embedding latent codes, enabling natural language search capabilities without the need for extra-label data. Splat-Nav~\cite{chen2024splat} proposes a real-time navigation system optimized for Gaussian Splat~\cite{kerbl20233d}-generated 3D scene representations. It consists of two key modules: Splat-Plan, which constructs collision-safe corridors and Bézier curve trajectories, and SplatLoc, facilitating robust pose estimation utilizing point cloud data and RGB images. Computational heavy-lifting, such as Bézier trajectory computation and pose optimization, is primarily handled by the CPU, allowing GPU resources to focus on tasks like online Gaussian Splat training. NeRF-IBVS~\cite{wang2024nerf} introduces a novel visual localization method aimed at achieving accurate localization with minimal posed images and 3D labels, addressing the challenge of acquiring such data in the real world. The method utilizes a coordinate regression network trained on a few posed images with coarse pseudo-3D labels from NeRF, followed by pose estimation with PnP and pose optimization with image-based visual servo (IBVS) leveraging scene priors from NeRF. NVINS~\cite{han2024nvins} introduces a novel framework that combines NeRF with Visual-Inertial Odometry (VIO) to enhance robotic navigation in real-time. By training an absolute pose regression network using augmented image data from NeRF and quantifying uncertainty, the approach addresses positional drift and improves system reliability. \mbox{Liu \textit{et al.}}~\cite{liu2024toward} introduces a navigation pipeline integrating NeRF into visuomotor navigation, emphasizing the importance of memory representations for intelligent agents. It presents a derivative radiance field for one-shot pose and depth estimation from a single query image, leveraging NeRF's spatial representation for task decomposition and action generation. Other applications of NFs for navigation include inventory monitoring that enables a mobile robot to continuously update its understanding of its environment~\cite{rashid2024lifelong}, visual localization~\cite{zhao2024pnerfloc} and exploration~\cite{camps2022learning}.

\subsubsection{Feature Field} Several works have looked into lifting 2D foundation features, {\em i.e.}, CLIP~\cite{radford2021learning}, DINO~\cite{oquab2023dinov2}, SAM~\cite{kirillov2023segment} for 3D scene understanding. This 3D distillation of 2D foundation models is trained per scene and it remains to be seen their generalizability to novel scenes or environment. Owing to the world knowledge present in 2D foundation models, 3D distillation enables many real-world applications, including mobile manipulation or navigation. 

\begin{figure}[t!] 
	\centering
	\includegraphics[width=1.0\columnwidth]{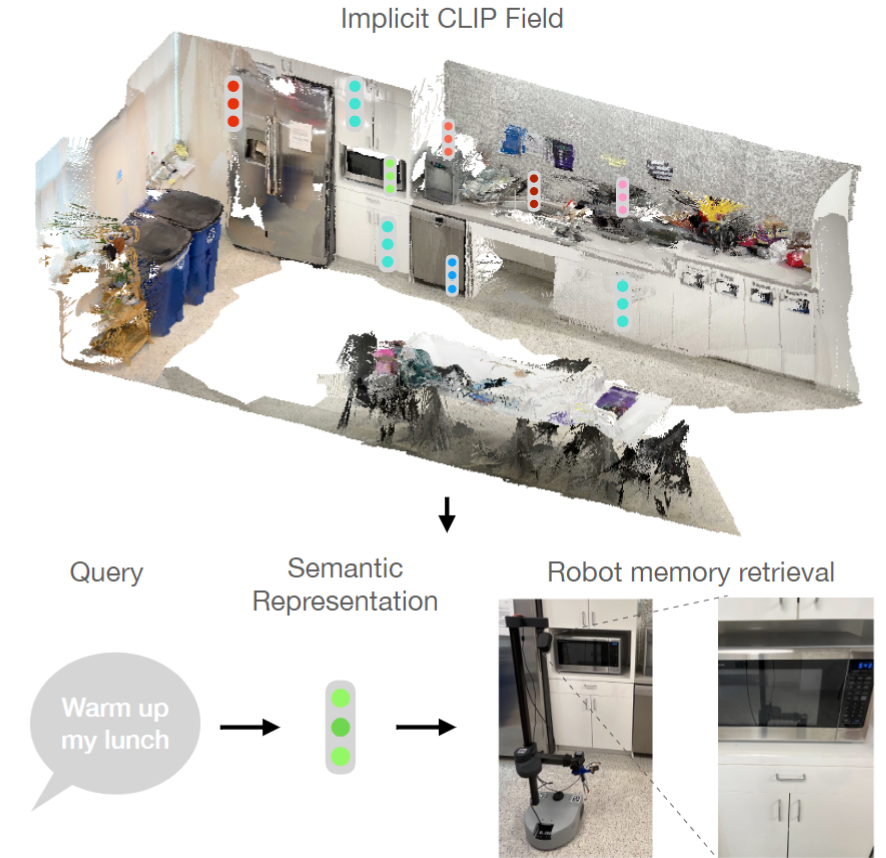}%
	\caption{Clip-Fields's~\cite{mahi2022clip} semantic representation enables 3D spatial memory for mobile robots.}
 \vspace{-0.5cm}
	\label{fig:clipfields}
\end{figure}

For mobile manipulation, explicit 2D feature distillation into 3D using pixel-aligned open-set features can be fused into 3D maps via traditional SLAM and multi-view fusion approaches~\cite{conceptfusion, huang23vlmaps}. CLIP-Fields~\cite{mahi2022clip}~(see Fig.~\ref{fig:clipfields}) implicitly utilizes a compact neural network to encode a 3D map and foundational features aligned with pixels or regions (such as LSeg~\cite{li2022language} and Detic~\cite{zhou2022detecting}). This specialized neural network, tailored to each scene, serves as a searchable database that aligns embeddings of images and language with 3D scene coordinates. It is designed to handle open-set queries specified in natural language. Language-embedded Radiance Fields (LEGS)~\cite{yu2024legs} extended LeRF~\cite{lerf2023} to train a queryable 3D representation online as the robot traverses the environment. It enables localization of open-vocabulary object queries while training significantly faster than LeRF. GaussNav~\cite{lei2024gaussnav} creates a map representation using 3D Gaussian Splatting. This framework allows the agent to remember both the geometric and semantic details of the scene, as well as the textural features of objects using MaskRCNN distillation into the 3D domain. 

Furthermore, Uni-fusion~\cite{yuan2024uni} proposes a universal continuous mapping framework for encoding surfaces and their properties ({\em e.g.}, color, infrared) without requiring training, using a Latent Implicit Map (LIM) that divides point clouds into voxels. It supports applications such as incremental reconstruction, 2D-to-3D property transfer, and open-vocabulary scene understanding. Open-Fusion~\cite{yamazaki2023open} proposes a real-time open-vocabulary 3D mapping and scene representation using RGB-D data, leveraging a pre-trained vision-language foundation model (VLFM) for semantic comprehension and the Truncated Signed Distance Function (TSDF) for rapid 3D reconstruction. It achieves annotation-free 3D segmentation without additional training and outperforms leading zero-shot methods.

Additionally, NF's compact 3D representation, as well as feature distillation, makes them ideal for integration with generative models~\cite{hong20233dllminjecting3dworld, hong2023threedclr}, where the obtained 3D features from NF's can be used directly in the projection space of the 2D Vision-Language models~\cite{alayrac2022flamingo, liu2023llava} to perform a diverse set of 3D related tasks such as 3D grounding, 3D visual question answering as well as navigation.

\subsubsection{Takeaways and Open Challenges in Neural Fields for Navigation}While Neural Fields have made significant strides in navigation, key challenges still remain. Current methods focus mainly on static environments and tasks like image-goal and vision-language navigation. Future work could extend NFs to dynamic settings, incorporating fast reconstruction techniques for real-time updates in evolving environments~\cite{puig2023habitat}. Another crucial direction is dynamic scene pose estimation~(\cref{key_takeaway_pose}) to aid reconstruction and navigation in dynamic environments.

The integration of generative NFs also holds great potential. Recent diffusion model advances~\cite{ho2020denoising, croitoru2023diffusion} could facilitate efficient scene editing and environment creation, narrowing the sim-to-real gap. Additionally, leveraging foundation models for large-scale mobile manipulation and scene generalization could unlock further advancements. Integrating Vision-Language Models (VLMs) with implicit representations for enhanced commonsense reasoning within NFs offers another promising frontier for future exploration.

\subsection{Neural Fields for Physics}
\label{sec:nf_physics}
Accurately simulating physics is a long-standing and challenging task traditionally marrying approaches from computer graphics and particle optimization. Linking these techniques with NFs opens up new possibilities, such as removing the need to explicitly model a scene while also imposing new challenges to researchers like balancing learned and non-learned parts. 

Given the novelty of the field, NFs have seen limited use in physics-based robotics applications. One notable example is ManiGaussian~\cite{lu2024manigaussian} (see \cref{subsec:manip:3D_gs}), though broader adoption remains sparse. In the following section, we discuss the possibilities and challenges introduced by the use of NFs in physical simulations for robotics. In \cref{sec:nf_physics:model_free}, we review model-free approaches that do not depend on explicit physical models. In contrast, \cref{sec:nf_physics:model_based} covers physically plausible, model-based methods.

\begin{figure}[t!]
    \centering
    \includegraphics[width=1.0\columnwidth]{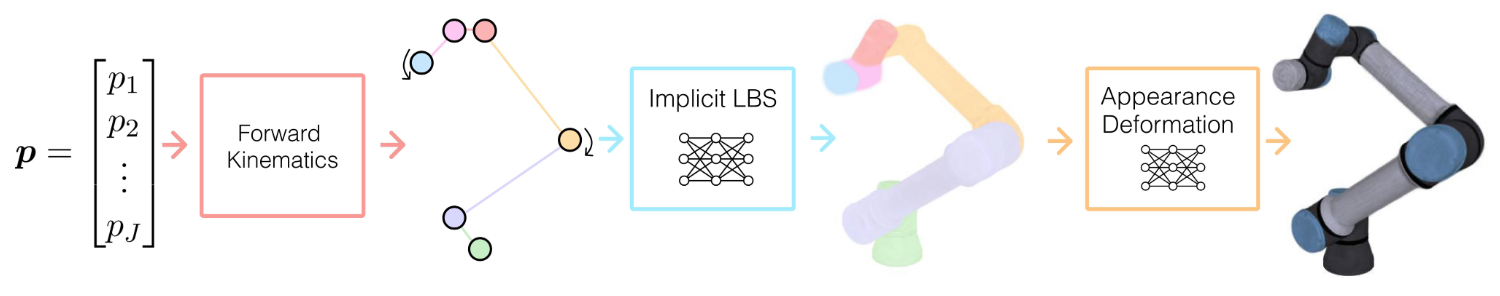}%
    \vspace{-0.5em}
    \caption{Differentiable Robot rendering pipeline~\cite{liu2024differentiable}.}
    \vspace{-0.4cm}
    \label{fig:drd_2024}
\end{figure}

\subsubsection{Model-Free}
\label{sec:nf_physics:model_free}
D-NeRF~\cite{pumarola2020d} was one of the first works that introduced a NeRF formulation that included a time component, allowing the representation of dynamic scenes. To decouple dynamics from structure, the authors learned an additional time-dependent MLP to map a spatial coordinate at a specific time step to a canonical space coordinate, which then serves as an input to a classical NeRF~\cite{park2021nerfies}. This technique is commonly known as a deformation field~\cite{liu2022devrf} and was extended to not just time but also arbitrary dimensions~\cite{park2021hypernerf}. 
Specifically, in the case of NeRFs, deformation fields were also coined as ray bending~\cite{tretschk2021non, qiao2022neuphysics}. 
Similarly, while \mbox{Li \textit{et al.}~\cite{li2021neural}} and \mbox{Gao \textit{et al.}~\cite{gao2021dynamic}} still use deformation fields to include the time component, both propose to explicitly regularize the reconstructed NeRF with an inferred scene flow.

A similar timeline of developments could be observed with 3D Gaussian Splatting. Here, \mbox{Luiten \textit{et al.}~\cite{luiten2023dynamic}} was one of the first to add a time-component to each Gaussian, influencing the 3D pose while keeping the size, color, and opacity constant. Consequently, applying deformation fields to enable dynamic Gaussian Splatting was first introduced by \mbox{Wu \textit{et al.}~\cite{wu20234d}}. \mbox{MD-Splatting~\cite{duisterhof2023md}} extended the approach to the metric space using a rigidity and isometry regularization term adapted from \mbox{Luiten \textit{et al.}~\cite{luiten2023dynamic}} and an additional momentum term. The extension to the metric space makes the reconstruction physically interpretable and thus allows applications in robotics. \mbox{Yang \textit{et al.}~\cite{yang2023gs4d}} take on a full probabilistic view and decompose the full 4D Gaussian (joint probability over space and time) into conditional Gaussian distributions. To reconstruct meshes, \mbox{Liu \textit{et. al}~\cite{liu2024dynamic}} introduce \mbox{DG-Mesh}, a method that utilizes a cycle-consistency loss on the predicted deformations. 

\begin{figure}[t!]
    \centering
    \includegraphics[width=1.0\columnwidth]{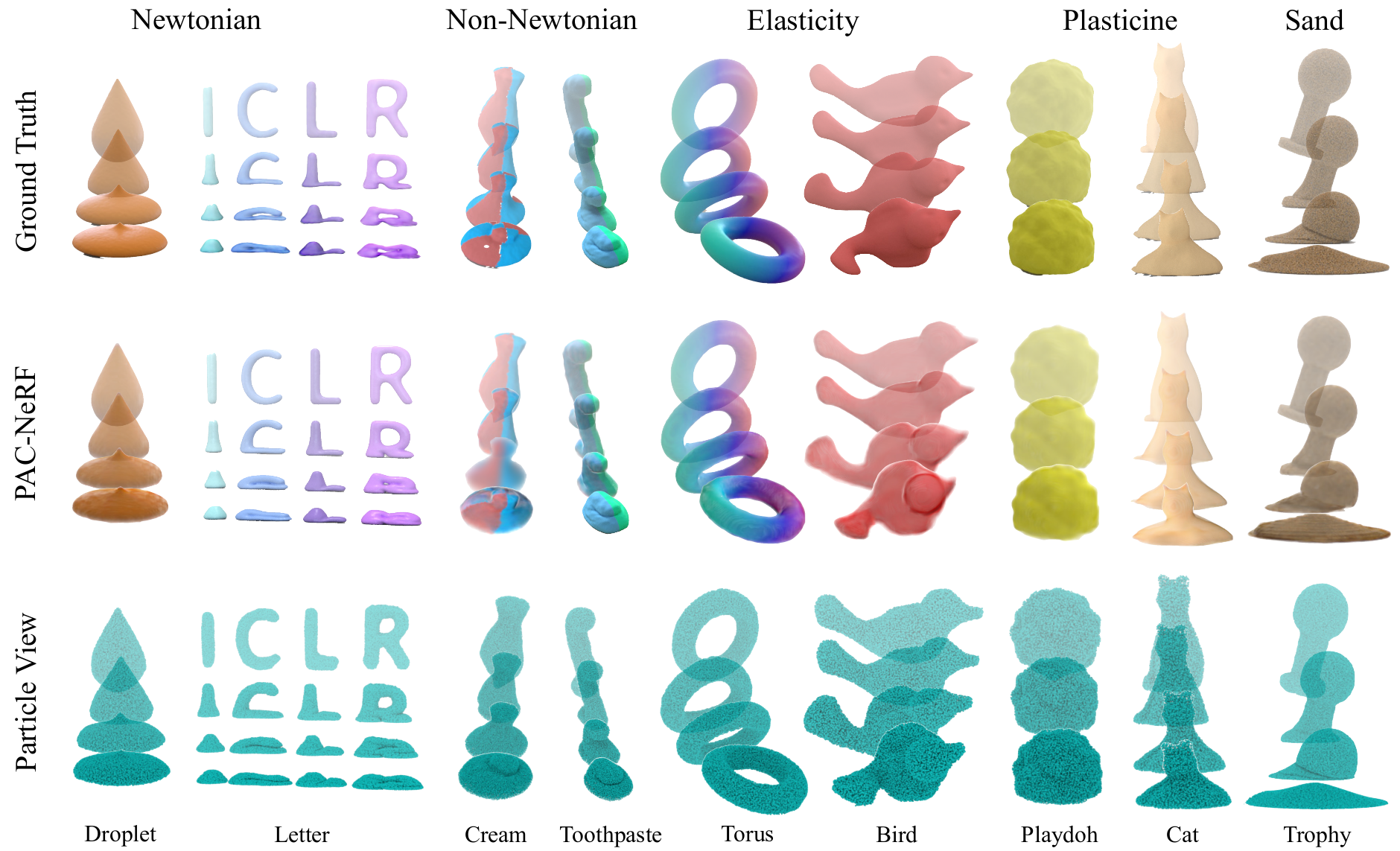}%
    \vspace{-0.5em}
    \caption{An overview of the different materials model-based NFs are able to simulate~\cite{li2023pacnerf}.}
    \vspace{-0.4cm}
    \label{fig:pac_nerf_materials}
\end{figure}

Instead of conditioning their NeRF on a time component, \mbox{Abou-Chakra \textit{et al.}~\cite{abou2024particlenerf}} approach the problem differently through particles that evolve over time and thus dynamically adapt to the scene in an online manner. They extended their approach also to Gaussian Splatting~\cite{abou2024physically} while also introducing the concept of visual forces. Nonetheless, a robotic input to interact with the scene is missing. Opposed to that most recently, \mbox{Li \textit{et al.}~\cite{li20223d}} proposed to first learn a latent state variable which then serves as an additional input to their NeRF. Given robot demonstrations, they then also learn a latent (implicit) dynamics model, inferring the next state given the current state and a robot action. Similarly, ManiGaussian~\cite{lu2024manigaussian} does not directly condition their Gaussians on a time component but rather on a robot action and thus, learn a Gaussian world model. Using known forward kinematics of a robot to condition Gaussians in Gaussian Splatting, \mbox{Liu \textit{et al.}~\cite{liu2024differentiable}} tackle the problem of differentiable robot rendering~(see Fig.~\ref{fig:drd_2024}), allowing to pass gradients from robot images down to the robot joint states, enabling various tasks such as text-to-robot pose, robot \& camera pose estimation, motion retargeting with point tracker and robot control with a generative video model~\cite{blattmann2023stable}.

To conclude, one could consider the aforementioned models as model-free as no underlying physics model is explicitly used nor enforced through regularization. 
While most of these methods presumably have a constant density~\cite{park2021nerfies, pumarola2020d}, it can not be guaranteed that the learning procedure will vanish parts through{\em, e.g.}, ray bending and thus produce physically implausible results. As a result, the utility of these models for reliable and safe robotics is unclear and yet to be explored.

\subsubsection{Model-Based}
\label{sec:nf_physics:model_based}
Opposed to model-free methods (see \cref{sec:nf_physics:model_free}), model-based methods incorporate underlying physical principles such as the aforementioned constant density~\cite{li2023pacnerf} and thus, could be considered as physically correct. In our analysis, we classify model-based methods based on their simulation scope: \textit{rigid}, \textit{articulated} and \textit{non-rigid}.

\looseness=-1
{\noindent \textbf{Rigid Objects}:} Similar to some of the aforementioned model-free methods, Hofherr \textit{et al.}~\cite{hofherr2023neural} used only a photometric loss for optimization, while constraining rigid object motion through an underlying physical dynamics model. Cleac’h \textit{et al.}~\cite{le2023differentiable} separated object reconstruction from parameter estimation, using static object images to train a neural representation and videos of moving objects to infer physical properties like friction and mass. Instead of videos, NeRF2Physics~\cite{zhai2024physical} distilled these properties from language and associated them with spatial language embeddings. MovingParts~\cite{yang2024movingparts} removed the assumption of object separation, thus automatically detecting coherently moving rigid parts and their transformations over time.

{\noindent \textbf{Articulated Objects}:} Articulated objects can be considered a class between purely rigid objects and non-rigid objects, as they impose constraints on rigid parts and their relative movement but do not decompose them down to the particle level. Most articulated object research has focused on pure visual reconstruction~\cite{heppert2023carto, tseng_cla-nerf_2022}, with limited emphasis on their physical interaction properties, such as manipulation and dynamics. Notably, some recent works have begun to bridge this gap by incorporating physical reasoning and interaction capabilities into models, allowing for a more comprehensive understanding and manipulation of articulated objects in real-world scenarios. Robot See Robot Do (RSRD)~\cite{kerr2024rsrd} enables robots to imitate articulated object manipulation from a single monocular video. RSRD introduces 4D Differentiable Part Models (4D-DPM) to reconstruct 3D part motion through an analysis-by-synthesis approach that optimizes geometric regularizes from a single video. This enables the robot to replicate object part motions by planning bi-manual arm trajectories, achieving notable success rates in physical execution without task-specific training or annotation.

{\noindent \textbf{Non-Rigid Objects}:} 
Non-rigid objects, unlike rigid and articulated ones, consist of numerous individual moving particles, making their simulation more complex yet generalizable. Non-rigid objects can be further categorized into subtypes such as deformable objects and fluids. One of the most versatile methods for simulating these objects is the Material Point Method (MPM), a framework capable of modeling a wide range of materials~\cite{li2023pacnerf, xie2023physgaussian}. PAC-NeRF~\cite{li2023pacnerf}~(see Fig. \ref{fig:pac_nerf_materials}) introduced a hybrid particle and grid-based NeRF representation that enables conversions between the two. In contrast, PhysGaussian~\cite{xie2023physgaussian} directly utilizes the particle-like nature of Gaussians in Gaussian Splatting, eliminating PAC-NeRF’s explicit conversion step. Similarly, PIE-NeRF~\cite{feng2023pie} avoids PAC-NeRF’s particle-to-rest-pose conversion, reducing over-smoothing. Spring-Gaus~\cite{zhong2024reconstruction} clusters Gaussians as point masses by inferring a static set of Gaussians followed by the sampling of anchor points ({\em i.e.}, particles).

To simulate fluids, Yu \textit{et al.}~\cite{yu2024inferring} split the velocity field into a base flow and a vortex particle flow, ensuring physical accuracy through a density and projection loss. Similar to PhysGaussian~\cite{xie2023physgaussian}, Gaussian Splashing~\cite{feng2024gaussian} uses Gaussian kernel centers as particles in a physics simulation. However, unlike PhysGaussian, the authors distinguish between solids and fluids, allowing them to first reconstruct a solid scene and then synthesize fluids within it. ClimateNerF~\cite{Li2023ClimateNeRF} followed a similar process, first reconstructing a scene with a classical NeRF pipeline before simulating different weather effects. Additionally, Zhong \textit{et al.}~\cite{zong2023neural} combined a neural deformation field with a Kirchhoff stress field in the same latent space, enabling faster, memory-efficient simulations by operating within this latent space.

\subsubsection{Takeaways and Open Challenges in Neural Fields for Physics}
Significant progress has been made in understanding and inferring physics; however, the challenge remains to seamlessly integrate these models with robots to create a truly simulatable, general, and interactive environment. Furthermore, it is still unclear how effectively policies learned in these simulations can be transferred to the real world.

\subsection{Neural Fields in Autonomous Driving}
\label{sec:nf_ad}
High-quality mapping of large-scale environments is essential for autonomous driving systems. A high-fidelity map of the entire operating domain serves as a powerful prior for various tasks, including robot localization~(see Sec.~\ref{sec:nf_pose}), navigation, and collision avoidance~(see Sec.~\ref{sec:nf_navigation}). Additionally, large-scale scene reconstructions facilitate closed-loop robotic simulations. Autonomous driving systems are often evaluated by re-simulating previously encountered scenarios; however, any deviation from the original encounter can alter the vehicle’s trajectory, necessitating high-fidelity novel view renderings along the adjusted path. In addition to basic view synthesis, scene-conditioned NeRFs can modify environmental lighting conditions, such as camera exposure, weather, or time of day, further enhancing simulation scenarios.

Neural Fields have become a prominent framework in autonomous driving due to their ability to generate photorealistic 3D environments from RGB images. These environments are highly valuable for constructing immersive simulation systems with several key features, as previously discussed: First, NFs offer extensive \textit{manipulability and compositionality}~(Sec.~\ref{subsec:manipulability}), allowing for the seamless integration and manipulation of objects within a scene. This facilitates the simulation of complex scenarios, such as collisions, which are difficult to replicate in physical settings. Second, they produce scenes with impressive \textit{photorealism}~(Sec.~\ref{subsec:simulators}), enabling realistic simulations from visual data. Finally, their strong \textit{generalizability}~(Sec.~\ref{subsec:generalizability}) from sparse inputs allows for creating accurate, scalable environments, enhancing research in embodied AI. These traits, as discussed in the following subsections, enable the creation of simulated environments that faithfully represent real-world scenarios, thereby facilitating research in embodied AI.

\subsubsection{Manipulability and Compositionality} \label{subsec:manipulability}
The underlying principle of utilizing photorealistic simulations lies in their efficacy as proxies for real-world environments in advancing research on embodied AI. Agents operating within these simulations can strategize and execute actions, thereby enhancing their ability to handle edge cases and facilitating smoother transitions to the real world with reduced domain gaps.

\begin{figure}[t!] 
	\centering
	\includegraphics[width=1.0\columnwidth]{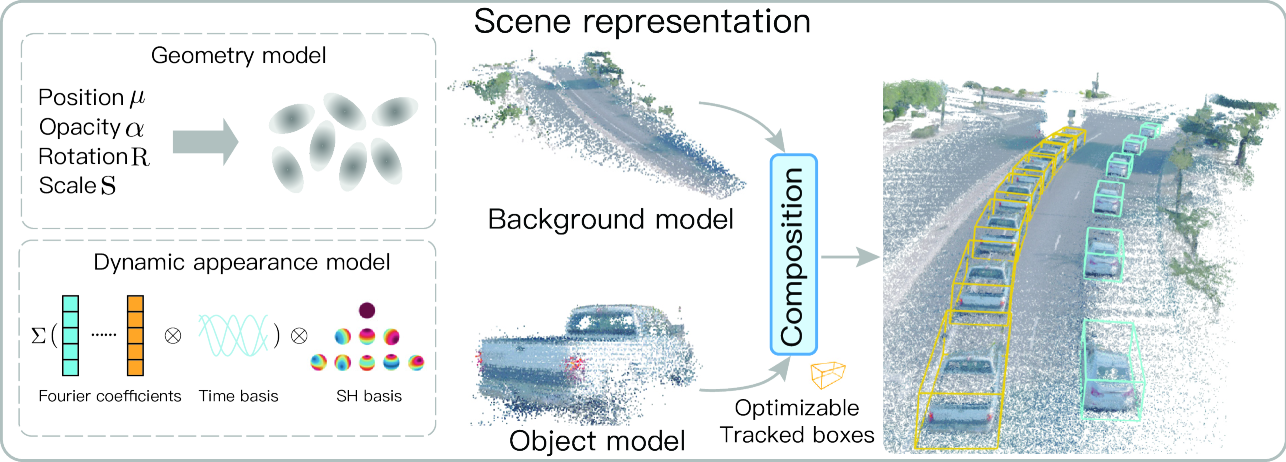}%
	\vspace{-0.5em}
	\caption{The compositional pipeline for Street Gaussians~\cite{yan2024street}.}
        \vspace{-0.3cm}
	\label{fig:streetgaussians}
\end{figure}

One of the first works that explored this paradigm is Neural Scene Graphs~\cite{ost2021neural}. It introduces a hierarchical approach to scene modeling, incorporating static and dynamic elements such as object appearance and shape. Utilizing a directed acyclic graph, scenes are uniquely defined, with nodes representing camera intrinsic, latent object codes and neural radiance fields for both static and dynamic elements and edges denoting transformations or property assignments. StreetNeRF~\cite{xie2023snerf} also focuses on compositional scene representation. It addresses limitations in traditional NeRFs for street-view synthesis by jointly considering large-scale background scenes and foreground moving vehicles. It improves scene parameterization and camera pose learning, leveraging noisy LiDAR points and geometry-based confidence to handle depth outliers. Experimental results demonstrate significant improvements in street-view synthesis and rendering moving vehicles compared to state-of-the-art methods. Similarly, Panoptic Neural Fields (PNF)~\cite{kundu2022panoptic} offers an object-aware neural scene representation, dividing scenes into objects and backgrounds. Leveraging compact MLPs for each object, PNF achieves faster processing while retaining category-specific priors, enabling tasks such as novel view synthesis, 2D panoptic segmentation, and 3D scene editing in real-world dynamic scenes. SUDS~\cite{turki2023suds} extends NeRFs for dynamic urban scenes by efficiently encoding static, dynamic, and far-field radiance fields using separate hash table data structures. Leveraging unlabeled target signals and various reconstruction losses, SUDS decomposes scenes into static backgrounds, individual objects, and their motions, achieving state-of-the-art performance on tasks like novel-view synthesis, unsupervised 3D instance segmentation, and 3D cuboid detection while significantly reducing training time compared to previous methods. EmerNeRF~\cite{yang2023emernerf} learns spatial-temporal representations of dynamic driving scenes by stratifying scenes into static and dynamic fields and parameterizing an induced flow field. Additionally, by lifting 2D visual foundation model features into 4D space-time, EmerNeRF improves semantic generalization and enhances 3D perception performance.

\begin{figure}[t!] 
	\centering
	 \vspace{-1em}
\includegraphics[width=1.0\columnwidth]{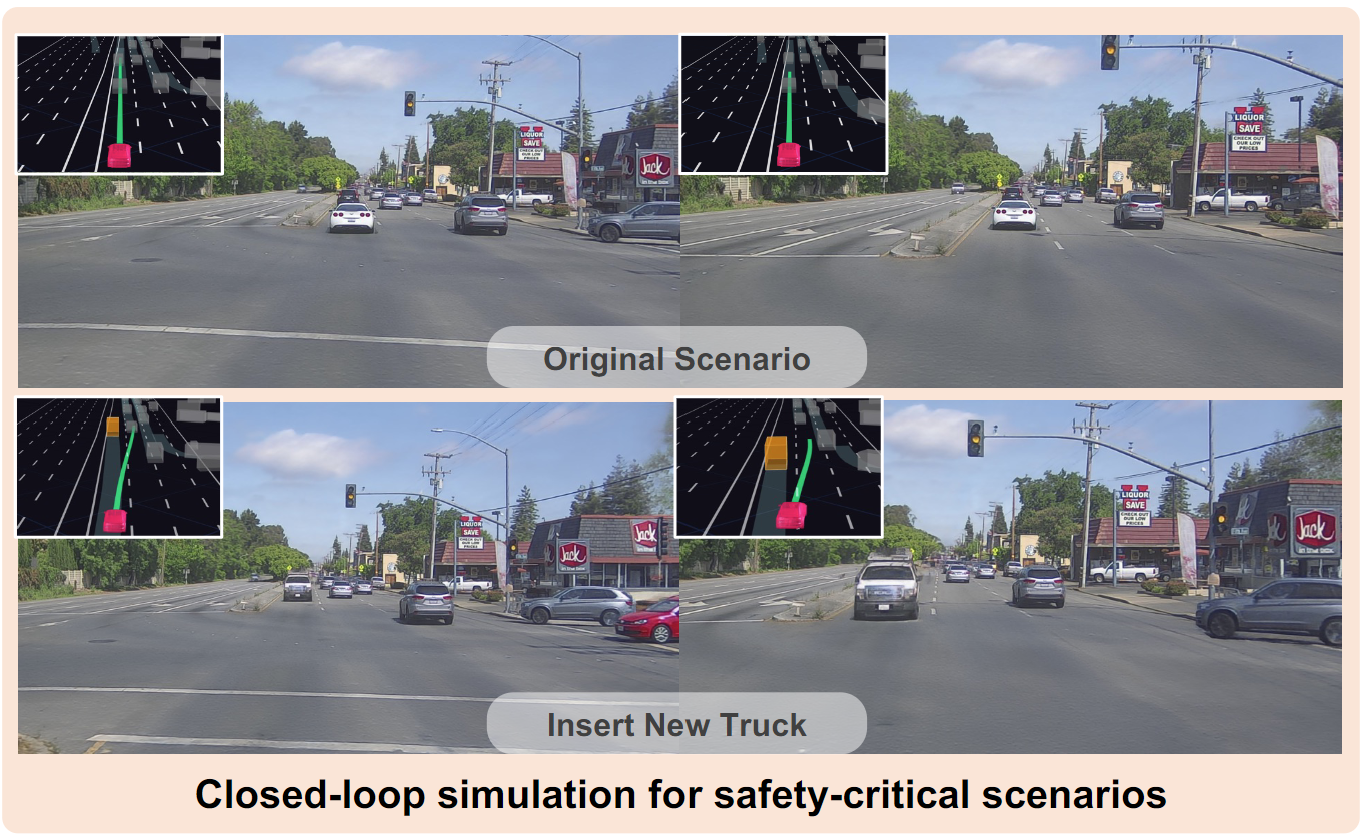}%
	\vspace{-0.5em}
	\caption{Photorealistic editing results from UniSim~\cite{yang2023unisim}.}
 \vspace{-0.5cm}
	\label{fig:unisim}
\end{figure}

Other works leverage explicit representations to accelerate rendering. Street Gaussians~\cite{yan2024street}~(see Fig.~\ref{fig:streetgaussians}) introduces a dynamic urban scene model using point clouds with semantic logits and 3D Gaussians for vehicles and backgrounds, enabling efficient scene editing and fast rendering. It outperforms prior methods on benchmarks using off-the-shelf tracker poses. Driving Gaussians~\cite{zhou2023drivinggaussian} reconstructs dynamic driving scenes with static 3D Gaussians and a dynamic Gaussian graph, using LiDAR priors for Gaussian splatting to achieve photorealistic synthesis and multi-camera consistency, surpassing existing techniques.

\subsubsection{Photorealistic Simulators} \label{subsec:simulators}
NeRFs excel in static scenes with controlled lighting conditions, but it faces difficulties when working with image collections from unpredictable real-world environments, which include varying weather, lighting, or temporary obstructions. NeRF-W~\cite{martinbrualla2020nerfw} seeks to overcome these challenges by using appearance embeddings and transient networks. Subsequent neural rendering for autonomous driving applications focused on building photorealistic simulators. MARS~\cite{wu2023mars} introduces an autonomous driving simulator based on NeRFs to address corner cases in autonomous vehicle driving. It features instance-aware modeling for separate foreground instances and background environments, proposing a modular design that enables flexible switching between different NeRF-related components, and achieves state-of-the-art photo-realism results. Notably, MARS is open-sourced, distinguishing it from most counterparts in the field. UniSim~\cite{yang2023unisim}~(see Fig.~\ref{fig:unisim}), a neural sensor simulator, enables closed-loop evaluation of self-driving vehicles by converting recorded driving logs into realistic multi-sensor simulations. Leveraging neural feature grids, UniSim reconstructs scene elements and dynamically simulates LiDAR and camera data, facilitating accurate assessment of autonomy systems on safety-critical scenarios. DriveEnv-NeRF~\cite{shen2024driveenv} proposes to use NeRFs to create high-fidelity simulations for validating and forecasting the performance of autonomous driving agents in real-world scenes. By rendering realistic images from novel viewpoints and constructing 3D meshes to emulate collisions, it bridges the sim-to-real gap, enhancing the robustness and real-world performance of autonomous driving agents compared to those trained with standard simulators. Lindström~{\em et al.}~\cite{lindstrom2024nerfs} propose methods to improve perception model robustness to NeRF artifacts, enhancing performance on both simulated and real data. Their large-scale investigation evaluates object detectors and an online mapping model on real and simulated data, demonstrating improved model robustness and, in some cases, better real-world performance.

\begin{figure}[t!] 
	\centering
\includegraphics[width=1.0\columnwidth]{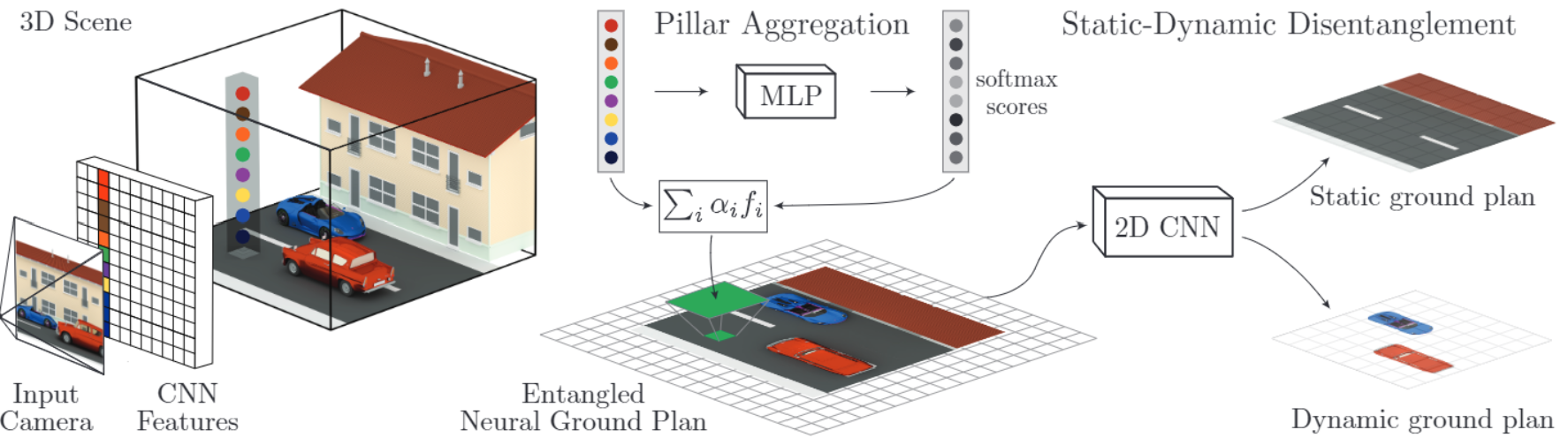}%
	\vspace{-0.5em}
	\caption{An overview of Neural Groundplans approach.~\cite{sharma2022neural}}
 \vspace{-0.5cm}
	\label{fig:neural_ground_plan}
\end{figure}

\subsubsection{Generalizability} \label{subsec:generalizability}
A separate line of works focused on generalizable systems for outdoor scenes. NeO 360~\cite{irshad2023neo360} introduces a novel approach for few-view view synthesis of outdoor scenes, overcoming limitations in existing methods by reconstructing 360$^{\circ}$ scenes from a single or few posed RGB images. By capturing the distribution of complex real-world outdoor 3D scenes and using a hybrid image-conditional tri-planar representation, NeO 360 offers generalizability to new views and novel scenes from as few as a single image during inference. 
Neural Groundplans~\cite{sharma2022neural}~(see Fig.~\ref{fig:neural_ground_plan}) introduces a method for mapping 2D image observations to a persistent 3D scene representation, facilitating novel view synthesis and disentangling movable and immovable scene components. Trained self-supervised from unlabeled multi-view observations, it leverages ground-aligned 2D feature grids inspired by bird's-eye-view representation, enabling efficient scene understanding tasks such as instance-level segmentation and 3D bounding box prediction. 6Imgto3D~\cite{gieruc20246imgto3d} uses a transformer-based encoder-renderer method designed for efficient and scalable single-shot 3D reconstruction from six outward-facing images in large-scale, unbounded outdoor driving scenarios. 

\subsubsection{Takeaways and Open Challenges in Neural Fields for Autonomous Driving}

Despite the promising progress in NFs for autonomous driving, several open challenges remain. Current methods focus on photorealistic simulators, which are dynamic, compositional, and realistic. One avenue of future work is training policies in such NF-based simulators and transferring them to the real-world. Connecting the success of NFs in autonomous driving with real-world deployment is an exciting avenue for future work. Generalizable reconstruction has seen some early signs of life with recent works but still remains largely underexplored. Future works could look at the efficiency of generalizable outdoor scene reconstruction methods, as well as advances that focus on sim2real transfer and pose-free reconstruction. This avenue of research is exciting as it opens the door for creating photorealistic simulators from a few images in the real world. 

Another promising direction for autonomous driving research is integrating generative methods like diffusion models with the NFs' paradigm.  Future work could look at creating new scenarios via NF editing that are difficult to create in the real world, such as collision avoidance to train policies via reward models in NFs' simulation. Generative asset creation through a few images from the real world is another potential avenue for NF's research for autonomous driving. 

Furthermore, the integration of NFs into generative models such as shown in Lift3D~\cite{li2023lift3d} and Adv3D~\cite{li2023adv3d} facilitates data augmentation, addressing the challenges posed by the diversity of driving scenes. Given the high costs associated with capturing all potential scenarios, data augmentation emerges as a valuable strategy and promising future direction for expanding training datasets and improving model performance.
\section{Open Challenges of Neural Fields in Robotics}
\label{sec:future_works}
Despite the exciting progress in the field, there are still several open challenges for various robotic applications to adopt Neural Fields. 
\begin{itemize}
    \item \textbf{Efficiency}: NFs are computationally intensive and may not naturally operate in real-time, which is often a critical requirement for robotics applications. There is a need for significant optimization or simplification to make these models run efficiently on robotics hardware, which may have limited computational resources compared to dedicated GPUs used in data centers.
    \item \textbf{Dynamic environments}: Robotics often involve operating in dynamic environments where objects and scene configurations change over time. Capturing and updating NFs to reflect these changes in real-time remains a challenging task.
    \item \textbf{Sensor integration}: Effectively integrating data from various sensors ({\em e.g.}, LiDAR, RGB cameras, depth sensors) to enhance the robustness and performance of NFs is relatively under-explored. Advanced sensor fusion techniques could potentially bridge this gap.
    \item \textbf{Generalization}: Existing techniques often require dense input data and struggle with sensor noise or occlusions. Developing methods that can leverage priors learned from web-scale datasets to generalize across varied scenarios offers a promising direction.
    \item \textbf{Physical information}: While NFs excel at representing visual aspects, they do not inherently understand physical properties like weight or friction. Extending NFs to incorporate physics simulations could enable more realistic interactions for robots.
    \item \textbf{Data efficiency and augmentation}: Current approaches are data-hungry, which is impractical for real-world applications. Innovations in data-efficient learning techniques and realistic data augmentation could help in overcoming these limitations.
    \item \textbf{Multi-modal, multi-task, and efficient scene understanding}: Developing neural field approaches that can handle multiple tasks and modalities simultaneously while maintaining efficiency in scene understanding is crucial for holistic robotic perception.
    \item \textbf{Performance evaluation}: Establishing standardized metrics and benchmarks for evaluating the performance of NFs in robotic applications will be essential for tracking progress and comparing different approaches.
    \item \textbf{Collaborative frameworks}: There is a need for frameworks that support collaboration between robots using NFs, enabling them to share learnings and improve collective understanding and decision-making in complex environments.
\end{itemize}

\ifCLASSOPTIONcaptionsoff
  \newpage
\fi

\renewcommand\refname{References}

{\footnotesize
\bibliographystyle{IEEEtran}
\bibliography{ref}
}

\end{document}